\documentclass{article}

\usepackage[preprint]{arxiv}

\usepackage[utf8]{inputenc} 
\usepackage[T1]{fontenc}    
\usepackage{hyperref}       
\usepackage{url}            
\usepackage{booktabs}       
\usepackage{amsfonts}       
\usepackage{nicefrac}       
\usepackage{microtype}      
\usepackage{xcolor}         

\usepackage{graphicx}
\usepackage{subfigure}
\usepackage{multirow}
\usepackage{multicol}
\usepackage{array}

\usepackage{amsmath}
\usepackage{amssymb}
\usepackage{mathtools}
\usepackage{amsthm}
\usepackage{subfigure}
\usepackage{subcaption}
\usepackage{wrapfig}

\RequirePackage{algorithm}
\RequirePackage{algorithmic}

\usepackage[capitalize,noabbrev]{cleveref}
\usepackage[most]{tcolorbox}

\title{Complex LLM Planning via Automated Heuristics Discovery}

\author{%
  Hongyi Ling\thanks{Equal contribution} 
  \quad 
  Shubham Parashar\footnotemark[1]
  \quad
  Sambhav Khurana\footnotemark[1]
  \quad
  \textbf{Blake Olson} \quad
  \textbf{Anwesha Basu} \\ 
  \textbf{Gaurangi Sinha} \quad 
  \textbf{Zhengzhong Tu} \quad
  \textbf{James Caverlee} \quad
  \textbf{Shuiwang Ji} \\ [5pt]
  Department of Computer Science \& Engineering\\
  Texas A\&M University \\[5pt]
}

\begin{document}

\maketitle

\begin{abstract}
We consider enhancing large language models (LLMs) for complex planning tasks. While existing methods allow LLMs to explore intermediate steps to make plans, they either depend on unreliable self-verification or external verifiers to evaluate these steps, which demand significant data and computations. 
Here, we propose automated heuristics discovery (AutoHD), a novel approach that enables LLMs to explicitly generate heuristic functions to guide inference-time search, allowing accurate evaluation of intermediate states. These heuristic functions are further refined through a heuristic evolution process, improving their robustness and effectiveness. Our proposed method requires no additional model training or fine-tuning, and the explicit definition of heuristic functions generated by the LLMs provides interpretability and insights into the reasoning process. Extensive experiments across diverse benchmarks demonstrate significant gains over multiple baselines, including nearly twice the accuracy on some datasets, establishing our approach as a reliable and interpretable solution for complex planning tasks.  Our code is released as part of the Sys2Bench library (\url{https://github.com/divelab/sys2bench/}).
\end{abstract}

\section{Introduction}




Large language models (LLMs) are increasingly being applied to tasks that require structured reasoning and decision-making, extending beyond traditional NLP applications such as translation and summarization. These models have demonstrated potential in complex reasoning tasks, including arithmetic problem-solving~\citep{cobbe2021gsm8k}, logical reasoning~\citep{PrOntoQA}, vision-based reasoning~\citep{parashar2024neglected}, and multi-step problem-solving~\citep{yang2018hotpotqa}. Building on these strengths, researchers have begun using LLMs for complex planning tasks~\citep{2025Sys2BenchLLM}, which require sequential decision-making, exploration of intermediate steps, and strategic thinking. 
To further enhance LLMs' capabilities in such tasks, test time inference techniques~\citep{wei2022chain,wang2022self,hao2023reasoning} like Tree-of-Thought~\citep{yao2024tree} enable LLMs to search over the intermediate steps, improving their ability to arrive at correct solutions. From solving puzzles to generating action plans and strategies, the use of LLMs for planning tasks continues to expand, showcasing their potential as reliable tools for tackling challenges that go beyond traditional NLP applications.

While test-time inference techniques have achieved notable success, there remain opportunities to improve them. Early approaches, such as Chain-of-Thought~\citep{wei2022chain}, rely on a linear reasoning process, which often fails to explore diverse solution paths or recover from errors introduced during intermediate steps. This limitation may result in suboptimal performance, particularly in complex planning tasks~\citep{valmeekam2022large} where early mistakes tend to propagate and result in incorrect final outcomes. More recent methods have attempted to address these challenges by incorporating mechanisms to search over intermediate steps and verify their correctness, using either LLM self-verification~\citep{hao2023reasoning} or external models to evaluate these steps~\citep{kambhampati2024llms}. However, self-verification has been shown to be unreliable, as LLMs cannot consistently identify errors~\citep{huang2024large}. On the other hand, approaches that require additional model training come with significant costs in terms of data and computational resources, making them less practical for many real-world applications.

Humans, on the other hand, rely on heuristics to simplify decision-making when faced with complex problems~\citep{hjeij2023brief,simon1997models,tversky1974judgment}. These heuristics serve as approximate rules that enable efficient problem decomposition and prioritization of actions, thereby reducing the need for exhaustive computation.
Inspired by this human problem-solving behavior, we propose automated heuristics discovery (AutoHD), a novel approach that addresses the limitations of existing methods. Specifically, AutoHD prompts LLMs to explicitly generate reliable heuristic functions represented as Python code. During inference, intermediate steps are evaluated using the generated heuristic function, guiding the search process effectively. To further enhance the robustness and performance of these heuristic functions, we introduce a heuristic evolution process that iteratively refines them over time. Notably, our approach achieves these advancements without requiring additional models or fine-tuning the pre-trained LLMs, making it lightweight and broadly applicable. Moreover, by explicitly defining the heuristic functions, AutoHD offers interpretability, allowing us to understand why LLMs prefer certain intermediate steps over others. Extensive experiments over multiple benchmarks, including Blocksworld, Rubik's Cube, and the Game of 24, demonstrate that AutoHD significantly improves the planning capabilities of LLMs, establishing a more reliable and interpretable framework for solving planning tasks.


\section{Background and Related Work}

We define a planning task specifically designed for solving with an LLM as $\{\mathcal{S}, s_0, \mathcal{G}, \mathcal{A}_\theta, T_\theta\}$. Here $\mathcal{S}$ represents the state space, containing all possible states. The initial state is denoted as $s_0 \in \mathcal{S}$, and the goal states are represented as $\mathcal{G} = \{ g_0,g_1, \cdots, g_m\}$, where each $g_i \in \mathcal{S}$ is a possible goal state. Note, in some cases, there may be only one goal state, i.e., $\mathcal{G} = \{ g_0\}$. The action space $\mathcal{A}_\theta(s)$ is generated by a pre-trained LLM with parameters $\theta$ based on the current state $s$, providing a set of valid actions for state $s$. The transition function $T_\theta : \mathcal{S} \times \mathcal{A} \rightarrow \mathcal{S}$, also parameterized by the LLM, predicts the outcome of applying an action $a \in \mathcal{A}_\theta(s)$ to a state $s$.  A solution to the planning task is represented as a state-action trace $\Pi = (s_0, a_1, s_1, a_2, \cdots, a_n,s_n )$ where $s_{i+1} = T_\theta(s_i, a_{i+1})$ and $s_n \in \mathcal{G}$.

\textbf{LLMs for Planning.} 
LLMs have demonstrated potential for reasoning and common-sense capabilities, driving their growing adoption in the planning tasks across various domains. Specifically, LLMs have been widely used as policy models for decision-making and planning in diverse interactive environments, including robotics~\citep{driess2023palm,huang2022inner,kannan2024smart,singh2023progprompt}, multimodal games~\citep{fan2022minedojo,wang2023voyager}, and text-based environments~\citep{liu2023agentbench,yao2022react,shinn2024reflexion}. To further enhance their performance on planning problems, researchers have developed test-time inference techniques that adaptively guide the model's reasoning and decision-making processes during inference time. For example, Chain of Thought (CoT) prompting~\citep{wei2022chain} guides LLMs to explicitly generate intermediate reasoning steps, yielding state-action traces for planning tasks. Building on this, Tree of Thoughts (ToT)~\citep{yao2024tree} and Graph of Thoughts (GoT)~\citep{besta2024graph} explore multiple paths to improve decision-making and robustness. Similarly, XoT~\citep{ding2023everything} trains an auxiliary policy model to assign rewards to solutions. Techniques like RAP~\citep{hao2023reasoning} utilize Monte Carlo Tree Search to navigate the expansive action space, while FoR~\citep{yu2024flow} introduces a fine-tuning approach to discover diverse and creative solutions across multiple planning tasks. Despite their effectiveness, these methods often rely on self-evaluation, which can be prone to reliability issues~\citep{huang2023large, stechly2024self}, or require additional models and fine-tuning, which entail significant computational overhead.


\textbf{LLMs for Automated Heuristic Discovery.} 
Automatic heuristic design refers to the process of generating, optimizing, or adapting heuristic functions automatically to solve problems. Traditional heuristic functions are typically handcrafted based on domain expertise, but automatic methods use algorithms to create or refine these heuristic functions without manual intervention. Evolutionary algorithms like genetic programming~\citep{koza1994genetic} have been widely used, allowing heuristic functions to evolve through processes inspired by natural selection. Recently, LLMs have emerged as powerful tools for code generation. Their extensive pretraining provides them with broad knowledge and contextual understanding, enabling them to assist in generating accurate code to solve various problems. For instance, \citet{liu2024large} uses LLMs as optimizers to directly generate new trial solutions via in-context learning. Similarly, FunSearch~\citep{romera2024mathematical} uses LLMs to guide evolutionary procedures in mathematical discovery. Furthermore, \citet{velivckovic2024amplifying} extend FunSearch by evolving scoring functions, significantly enhancing performance on combinatorial competitive programming tasks. \citet{liu2024evolution} propose to use LLM with evolutionary computation methods for automatic heuristic design. However, automated heuristic discovery for LLM planning remains largely unexplored, and our work is the first to investigate this.

\begin{figure}[t]
    \begin{center}
    \begin{tabular}{@{\hspace{0cm}}c@{\hspace{-0.45cm}}c@{\hspace{-0.45cm}}c@{\hspace{-0.45cm}}c}
    \includegraphics[height=5.2cm]{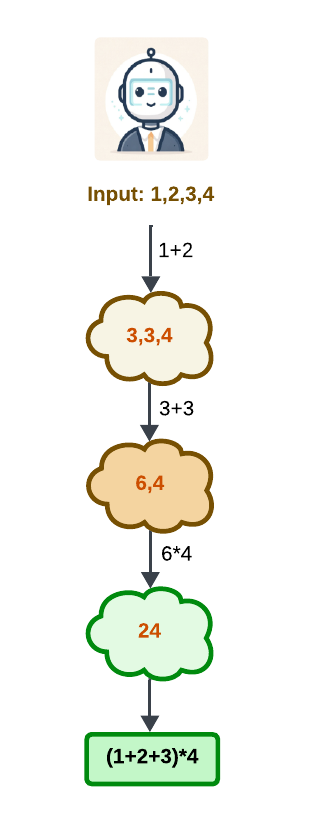} &
    \includegraphics[height=5.2cm]{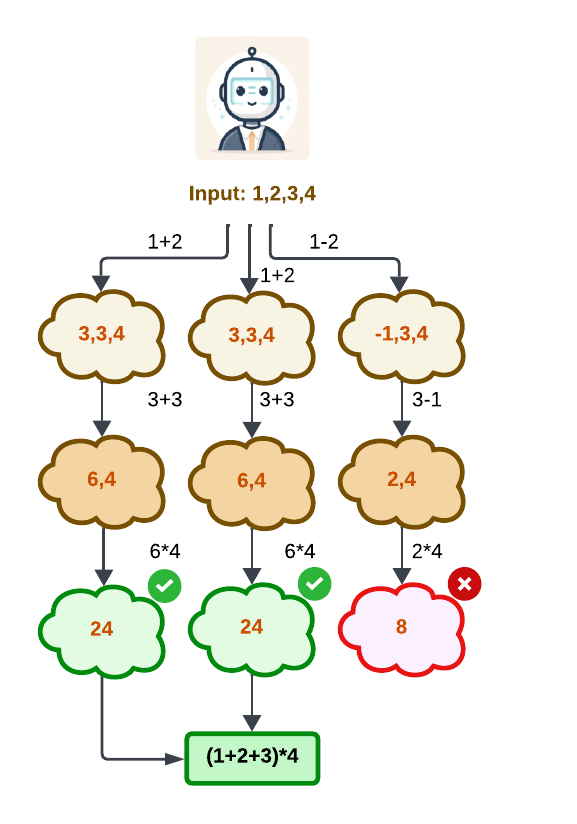} &
    \includegraphics[height=5.2cm]{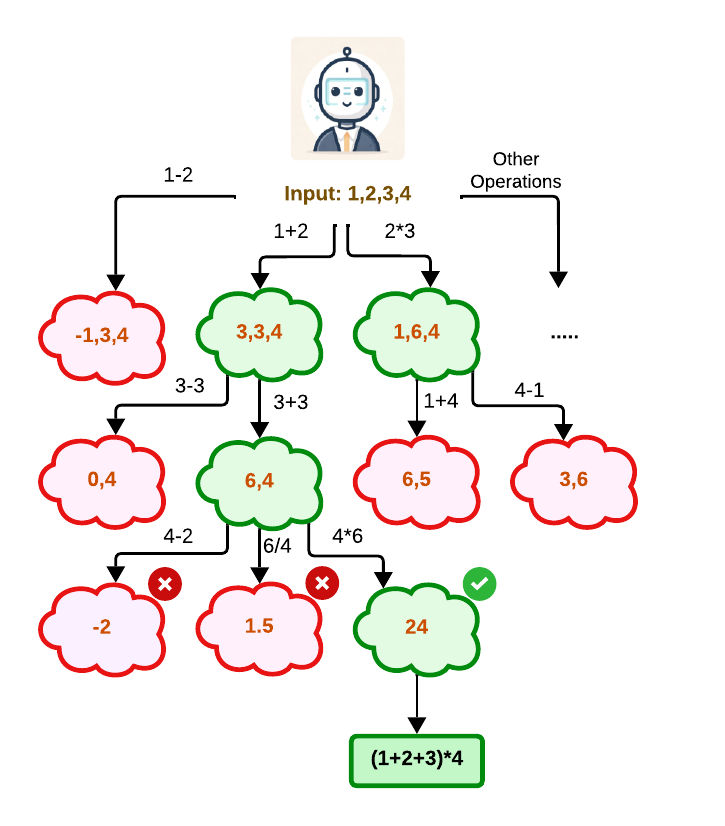} & 
    \includegraphics[height=5.2cm]{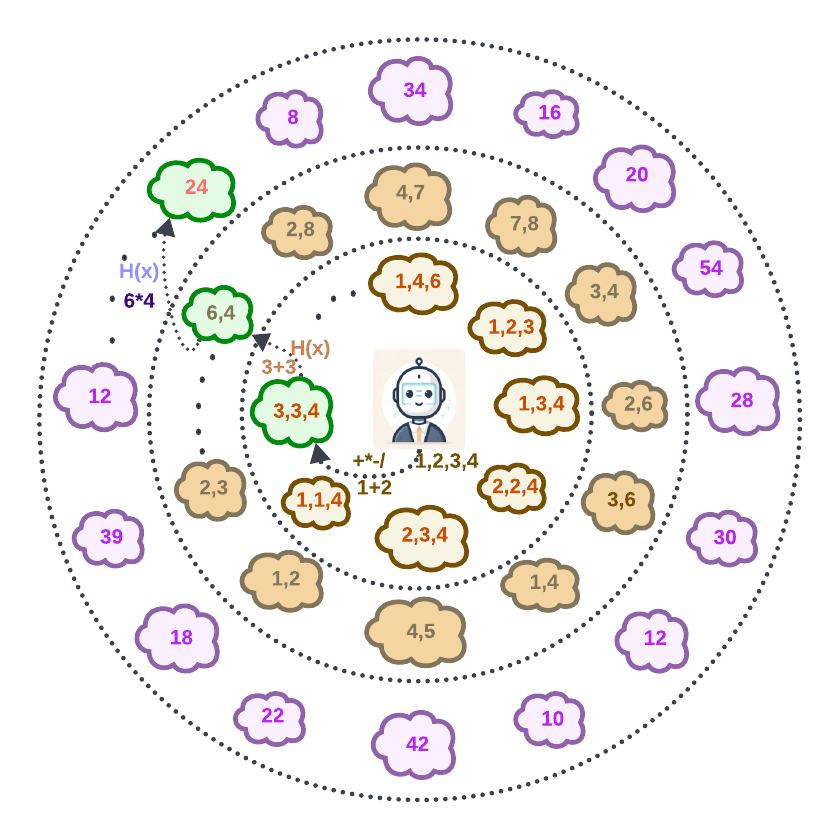} \vspace{-0.0cm} \\
    (a) CoT & (b) CoT-SC & (c) ToT & (d) AutoHD \\
    \end{tabular}
    \end{center}\vspace{-0.0cm}
    \caption{\small Comparison between existing methods and the proposed method. (a) CoT follows a single linear path, thereby constraining its exploratory capacity. (b) CoT-SC extends this approach by performing multiple CoT iterations, leading to a result with higher confidence scores. (c) ToT introduces a tree-based search mechanism, branching systematically through intermediate states to explore a broader solution space. (D) In contrast, our AutoHD uses a heuristic function generated by the LLM to guide exploration. The heuristic prioritizes promising states, enabling more efficient and effective navigation of the solution space. }
     \label{fig:comparison}
     \vspace{-0.0cm}
\end{figure}

\section{Automated Heuristics Discovery}
Existing methods rely on LLMs to either self-verify or use additional models to evaluate the quality of each intermediate state $s_i$~\citep{yao2024tree,ding2023everything}. However, prior research~\citep{huang2023large, stechly2024self} has shown that self-verification is often unreliable. Meanwhile, additional model training requires significant costs in terms of data and computational resources. 
To overcome these challenges, we propose AutoHD, a novel approach that enables LLMs to explicitly generate heuristic functions to guide inference-time search. Using the exceptional code generation capabilities of LLMs, AutoHD generates heuristic functions as Python code, enabling more accurate evaluation of intermediate states. To further improve the performance and robustness of these heuristic functions, we introduce a heuristic evolution process that iteratively refines them. Our framework reduces the inherent randomness of self-verification and enhances interpretability by providing explicit reasoning for why LLMs prioritize certain states. See Figure~\ref{fig:comparison} for a comparison between our proposed method and existing approaches.

\subsection{Heuristic Function Proposal}
A diverse set of initial heuristic functions is generated by prompting a pre-trained LLM, offering a flexible and automated solution that avoids reliance on hand-crafted components or additional models. The LLM generates both a natural language description and the corresponding Python code for each heuristic. The natural language description provides an intuitive, high-level overview of the heuristic function, while the Python code specifies its detailed implementation. These generated heuristic functions are designed to evaluate the quality or proximity of a given state $s$ relative to the goal states $\mathcal{G}$. 
\begin{wrapfigure}{r}{0.5\textwidth}
    \vspace{-0.0cm}
    \includegraphics[width=1.0\linewidth]{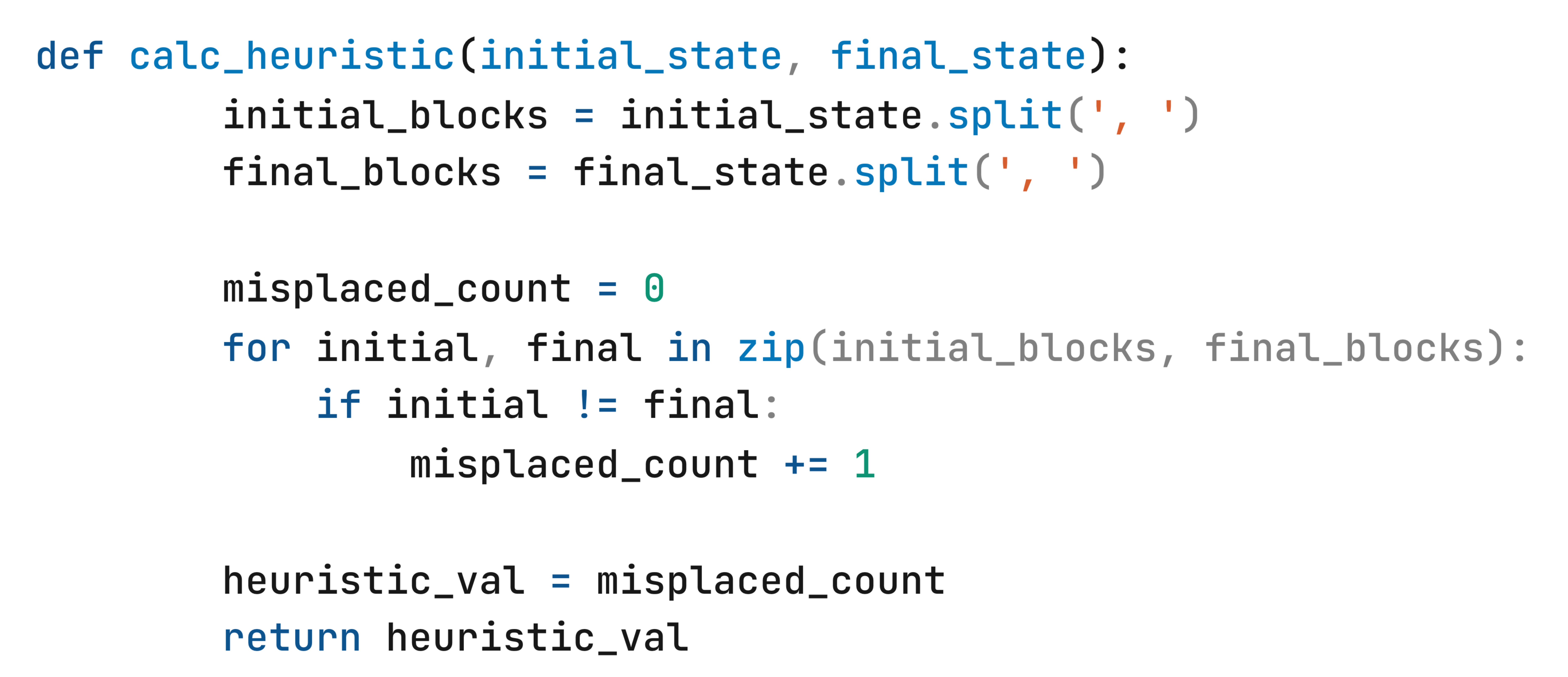}
    \caption{ An example heuristic function proposed by LLMs for Blocksworld. The heuristic function computes the number of misplaced blocks and the cumulative positional differences of these blocks. The resulting sum provides an estimation of the discrepancy between states.}
    \label{fig:heuristic}
    \vspace{-0.4cm}
\end{wrapfigure}
We denote the generated heuristic function as $H \in \mathcal{H}:\mathcal{S} \times 2^{\mathcal{S}} \rightarrow \mathbb{R}$, where $2^{\mathcal{S}}$ indicates the power set of the state space $\mathcal{S}$.
An example is provided in Figure~\ref{fig:heuristic}, which demonstrates a heuristic function for the Blocks World problem. The heuristic function estimates the cost of transitioning from an initial to a goal state in the Blocks World problem. It counts the number of blocks not in their correct positions, capturing structural misalignment. Besides, it computes the cumulative absolute positional difference of misplaced blocks, reflecting the effort required to reorder them. By summing these two, the heuristic provides a scalable and interpretable estimate of the discrepancy between states. See Appendix~\ref{appendix:heuristic_proposal_prompt} for the detailed prompt used in our experiments.

\subsection{Heuristic Guided Inference-time Search} \label{subsec:search}

Given an LLM-proposed heuristic function $H$, it efficiently guides the search process. This heuristic function determines the order in which states are explored, ensuring that the search algorithm prioritizes promising paths while pruning unimportant ones. In this way, the heuristic plays a crucial role in balancing between exploring new states and exploiting known promising paths. 
Specifically, at a state $s$, the LLM generates the corresponding action space $\mathcal{A}_\theta(s)$, which contains all feasible actions from that state. For a given action $a \in \mathcal{A}_\theta(s)$, the LLM predicts the resulting next state $s'$ as $s' = T_\theta(s, a)$. The proposed heuristic function $H$ then computes a heuristic value $v$ for the predicted state $s'$, providing an estimation of desirability. Formally, the heuristic values for all possible next states are computed as $$
v_i = H(T_\theta(s, a_i)),\quad \text{for} \ a_i \in \mathcal{A}_\theta(s).
$$
These values, collectively represented as $\mathcal{V} = \{v_1,v_2,\cdots\}$, guide the search algorithm by prioritizing the exploration of states with lower heuristic values. The heuristic values can be integrated into various search algorithms, such as $\text{A}^*$, beam search, or greedy search, to guide the exploration process. These values provide an estimation of the potential value of states, enabling the algorithms to prioritize or prune states dynamically and adapt to the specific requirements of the search strategy.
In our work, we explore two search algorithms below.

\textbf{Greedy Breadth-First Search.} 
Greedy breadth-first search (Greedy BFS) is a heuristic-driven search algorithm that combines the exploration of breadth-first search with greedy prioritization. At each step, it evaluates all states in the frontier, which is the set of all states that have been generated but not yet explored or expanded, and selects the one with the smallest heuristic value. In this way, Greedy BFS effectively chooses the most promising states at each step. 
Let $Q$ represent the frontier set at a given step. For each state $s' \in Q$, the algorithm evaluates the heuristic value $H(s')$. It then selects the next state to expand as $s = \text{argmin}_{s' \in Q} H(s')$, where $H(s')$ is the heuristic value of state $s'$. The algorithm terminates when a goal state is reached or the search budget is exhausted.
See Algorithm~\ref{alg:bfs} for details. 

\textbf{$\text{A}^*$ Search.} $\text{A}^*$ search is a widely used algorithm that balances exploration and exploitation using a composite cost function. It evaluates states based on the cumulative cost to reach the state $G(s)$, and the estimated cost to reach the goal $H(s)$. In our case, each step has a uniform cost, meaning the cumulative cost $G(s)$ represents the number of steps from the start state $s_0$ to the state $s$. Formally, for each step, $\text{A}^*$ maintains a priority queue of frontier nodes, which is the set of all states that have been generated but not yet expanded, sorted by the summed cost $F(s) = G(s) + H(s)$.  At each step, the algorithm selects the node $s$ with the smallest $F(s)$ value as $s = \text{argmin}_{s' \in Q} F(s')$. The search continues until a goal state is expanded or the search budget is reached. See Algorithm~\ref{alg:astar} for details. 
\begin{figure}[t]
    \centering
    \includegraphics[width=1.0\linewidth]{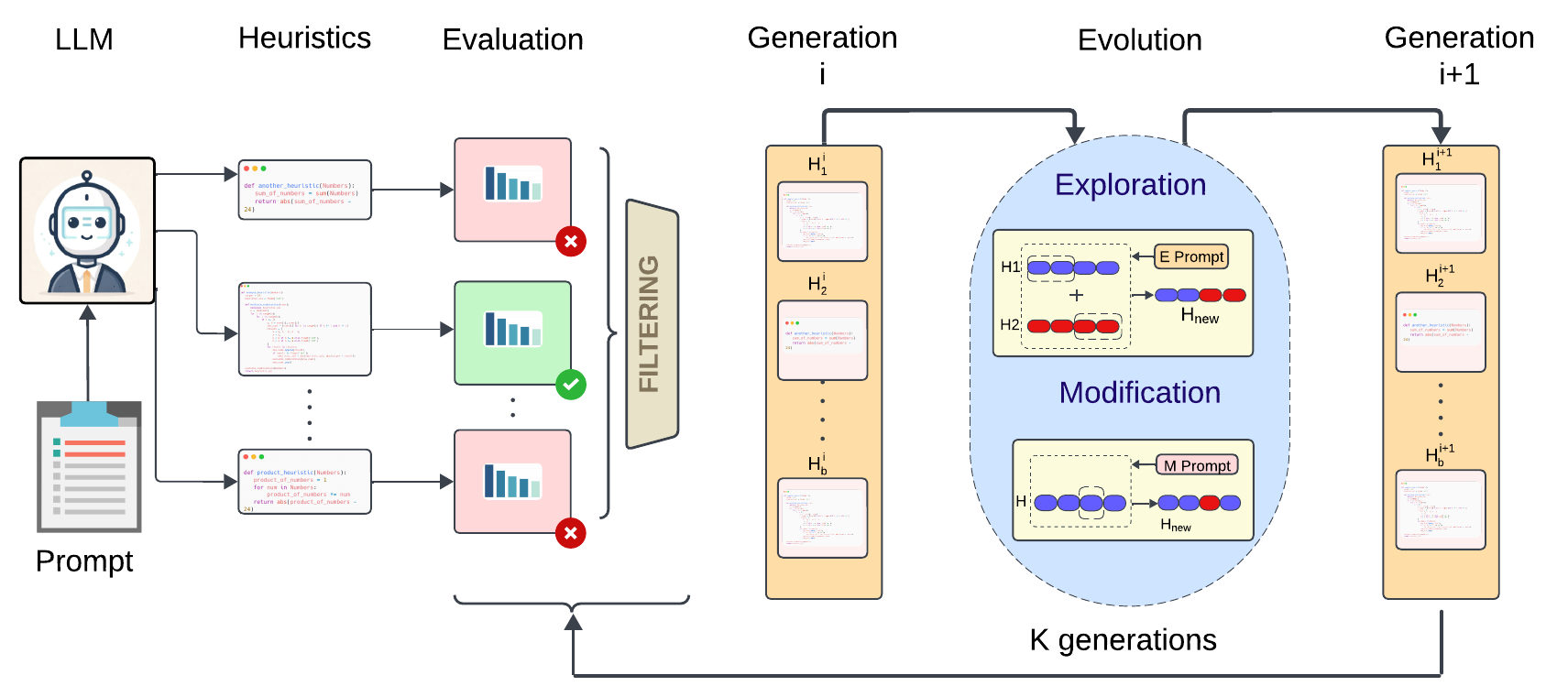}
    \vspace{-0.6cm}
    \caption{\small Heurisitc discovery process of the proposed method. The LLM is prompted to generate a diverse set of initial heuristic functions. These functions are evaluated on validation sets through heuristic-guided search to assess their quality. The top-performing heuristic functions are filtered and evolved to create the next generation by exploring new heuristic functions and refining existing ones. After K evolutions, the best heuristic function across all generations is selected for testing.}
    \label{fig:overview}
    \vspace{-0.4cm}
\end{figure}

\subsection{Heuristic Evolutions}


To further enhance the performance and robustness of the proposed heuristic functions, we follow~\citet{liu2024evolution,romera2024mathematical} to include a heuristic evolution mechanism. In this approach, LLMs are prompted to generate initial heuristic functions, and subsequent generations of heuristic functions are derived iteratively. At each generation, new heuristic functions are evaluated, and only the top-performing ones are retained for the next round. Specifically, LLMs are prompted to generate an initial pool of $b$ heuristic functions, denoted as ${H_1^{0}, H_2^{0}, \cdots, H_b^{0}}$. The detailed prompts for each problem are provided in Appendix~\ref{appendix:heuristic_proposal_prompt}. A small validation set consisting of approximately 10 problem instances, similar in size to the in-context learning examples, is used to evaluate the quality of each heuristic function. In generation $i$, given the heuristic functions ${H_1^{i-1}, H_2^{i-1}, \cdots, H_b^{i-1}}$ from the previous generation, we use two strategies for generating new heuristic functions, including exploration and modification. The exploration strategy focuses on generating new heuristic functions, encouraging the exploration of diverse ideas. On the other hand, the modification strategy introduces minor variations to existing high-performing heuristic functions, such as changing parameters, to refine and improve them. This combination of strategies ensures both comprehensive exploration and efficient refinement of the heuristic function space. Details of the evolution prompts are provided in Appendix~\ref{appendix:heuristic_evolution_prompt}. The newly generated heuristic functions ${H_1^{i}, H_2^{i}, \cdots, H_b^{i}}$ are then evaluated on the validation set through the heuristic-guided search process. The top-performing heuristic functions are selected to form the pool for the next generation. After several rounds of evolution, the best heuristic function across all generations is selected for testing. This process is performed only once, and no additional heuristic function generations are required during the inference-time search. The heuristic evolution process is summarized in Algorithm~\ref{alg:he}. See Figure~\ref{fig:overview} for an overview of the heuristic discovery process.

\begin{algorithm}
    \caption{Heuristic Evolution}\label{alg:he}
\begin{algorithmic}
\STATE Initialize a pool of $b$ heuristic functions $\mathbb{H}^0 = \{H_1^{0}, H_2^{0}, \cdots, H_b^{0}\}$ 
\FOR{$i \leftarrow 1$ to $K$}
\STATE Initialize an empty set $\mathbb{H}^{i} = \emptyset$
\STATE $\mathbb{H}^{i} = \mathbb{H}^{i} \cup \text{Exploration}(\mathbb{H}^{i-1})$
\STATE $\mathbb{H}^{i} = \mathbb{H}^{i} \cup \text{Modification}(\mathbb{H}^{i-1})$
\STATE Evaluate each heuristic function $H \in \mathbb{H}^{i}$ on the validation set using heuristic-guided search
\STATE $\mathbb{H}^{i} \gets \text{sort}(\mathbb{H}^{i}, \text{validation accuracy)}$
\STATE $\mathbb{H}^{i} \gets \mathbb{H}^{i}[:\lfloor |\mathbb{H}^{i}| / 2 \rfloor]$
\STATE $\mathbb{H}^{i} \gets \text{sample}(\mathbb{H}^{i}, b)$
\ENDFOR
\end{algorithmic}
\end{algorithm}

\subsection{Discussions}


Unlike existing methods that rely on LLM self-verification~\citep{yao2024tree, hao2023reasoning} or guidance from external models~\citep{kambhampati2024llms}, our approach uses LLMs to propose heuristic functions explicitly, which can guide the search process during inference. 
Additionally, while ToT requires invoking LLMs to evaluate at every intermediate step during inference, our framework eliminates this overhead. Once the best heuristic function is identified on the validation set, no further heuristic function generation is necessary during inference. See Figure~\ref{fig:comparison} for a comparison between our proposed method and existing approaches.

Previous works point out that an LLM can itself be treated as a heuristic. For instance, ToT~\citep{yao2024tree} considers its own framework as a heuristic search algorithm, where the LLM serves as the heuristic by evaluating and ranking nodes during the search. However, prior methods~\citep{stechly2024self,huang2023large} demonstrate that LLMs are inherently unable to self-verify, making them unreliable for providing robust heuristic values. 
Additionally, using the LLM itself to evaluate states lacks interpretability. This is because the LLM acts as a ``black-box" evaluator, providing no insight into why certain states are preferred over others. As a result, the reasoning process becomes obscured, making it difficult to understand the rationale behind the LLM's decisions. 
In contrast, AutoHD enhances transparency by asking the LLM to generate these heuristic functions explicitly. This approach provides insight into the decision-making process, enabling a clearer understanding of why specific states are prioritized.

Additionally, some studies have explored process reward models (PRMs)~\citep{luo2024improve, li2022making, lightman2023let}, which enhance LLMs by evaluating and optimizing intermediate steps of reasoning or decision-making processes. It is worth noting that in planning tasks, heuristic functions and PRMs serve a similar purpose, as both can guide the search. However, training PRMs often demands a large and diverse dataset of labeled intermediate states and their corresponding rewards, which can be challenging and costly to acquire, particularly for tasks with sparse data availability or where domain expertise is required to label the data~\citep{lightman2023let}. In contrast, AutoHD shows that LLMs can generate heuristic functions on the fly, taking advantage of their pre-trained knowledge without requiring additional training or manual data collection. This efficiency makes LLM-proposed heuristic functions a desirable alternative to traditional PRMs in many applications. 

\begin{table*}[t]
\caption{An overview of datasets including Blocksworld, Game of 24, Rubik's Cube.}
\label{tab:dataset-desc}
\centering
\begin{tabular}{@{}llll@{}}
\toprule
 & Blocksworld & Game of 24 & Rubik's Cube \\ \midrule
Task & \begin{tabular}[c]{@{}l@{}}Propose actions to \\ transform an initial block\\ configuration to a goal \end{tabular} & \begin{tabular}[c]{@{}l@{}}Use a list of four numbers\\ to make 24, through \\ +, -, $\times$ and /\end{tabular} & \begin{tabular}[c]{@{}l@{}}Rotate a $2\times2$ Rubik's \\ cube \includegraphics[width=.02\linewidth]{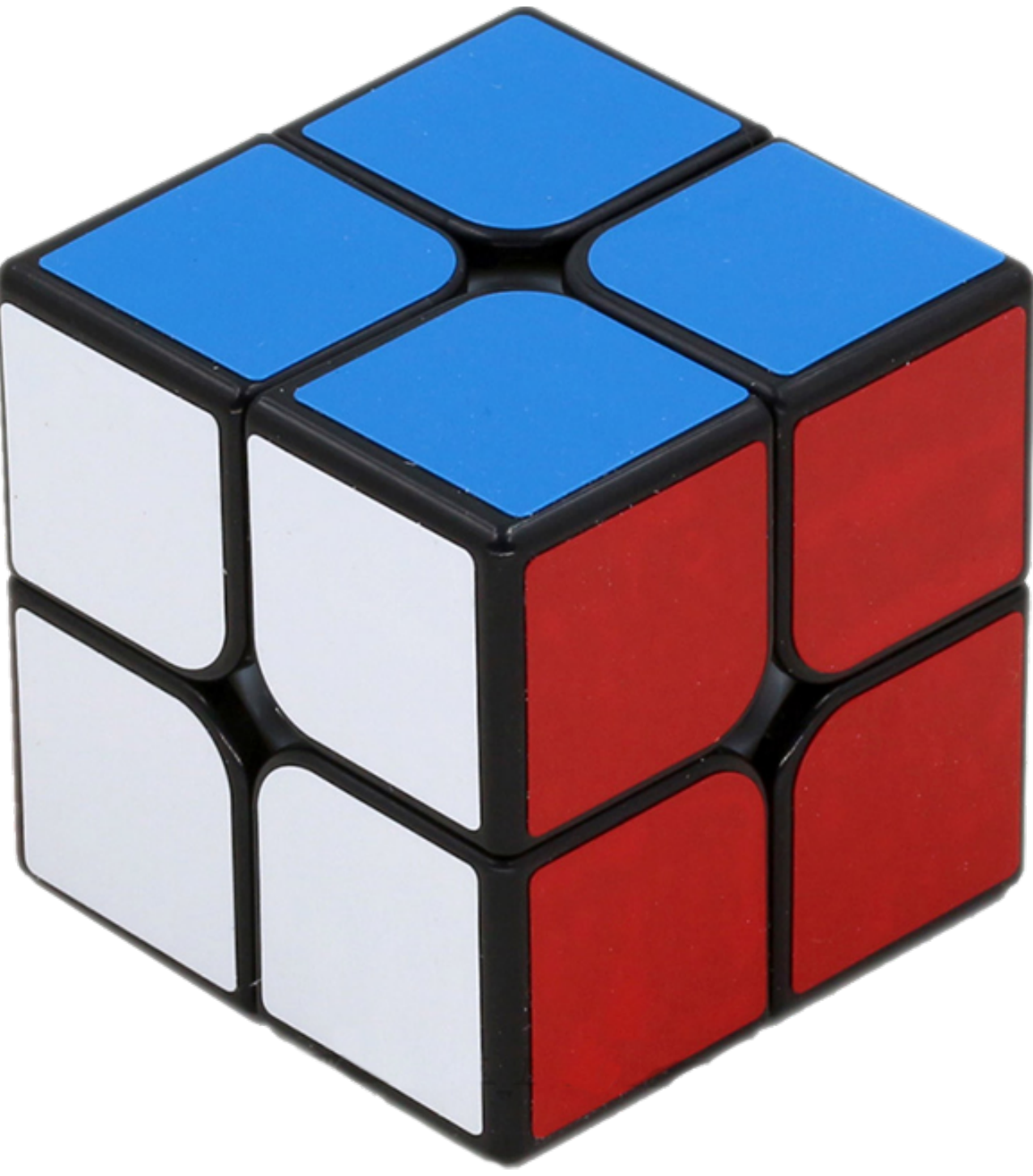} until each face \\ has one single color \end{tabular}\\ \midrule
Input & \begin{tabular}[c]{@{}l@{}}A starting configuration \\ of blocks\end{tabular} & \begin{tabular}[c]{@{}l@{}}A starting list of \\ 4 numbers\end{tabular} & \begin{tabular}[c]{@{}l@{}}A scrambled $2\times2$ \\ Rubik's cube\includegraphics[width=.02\linewidth]{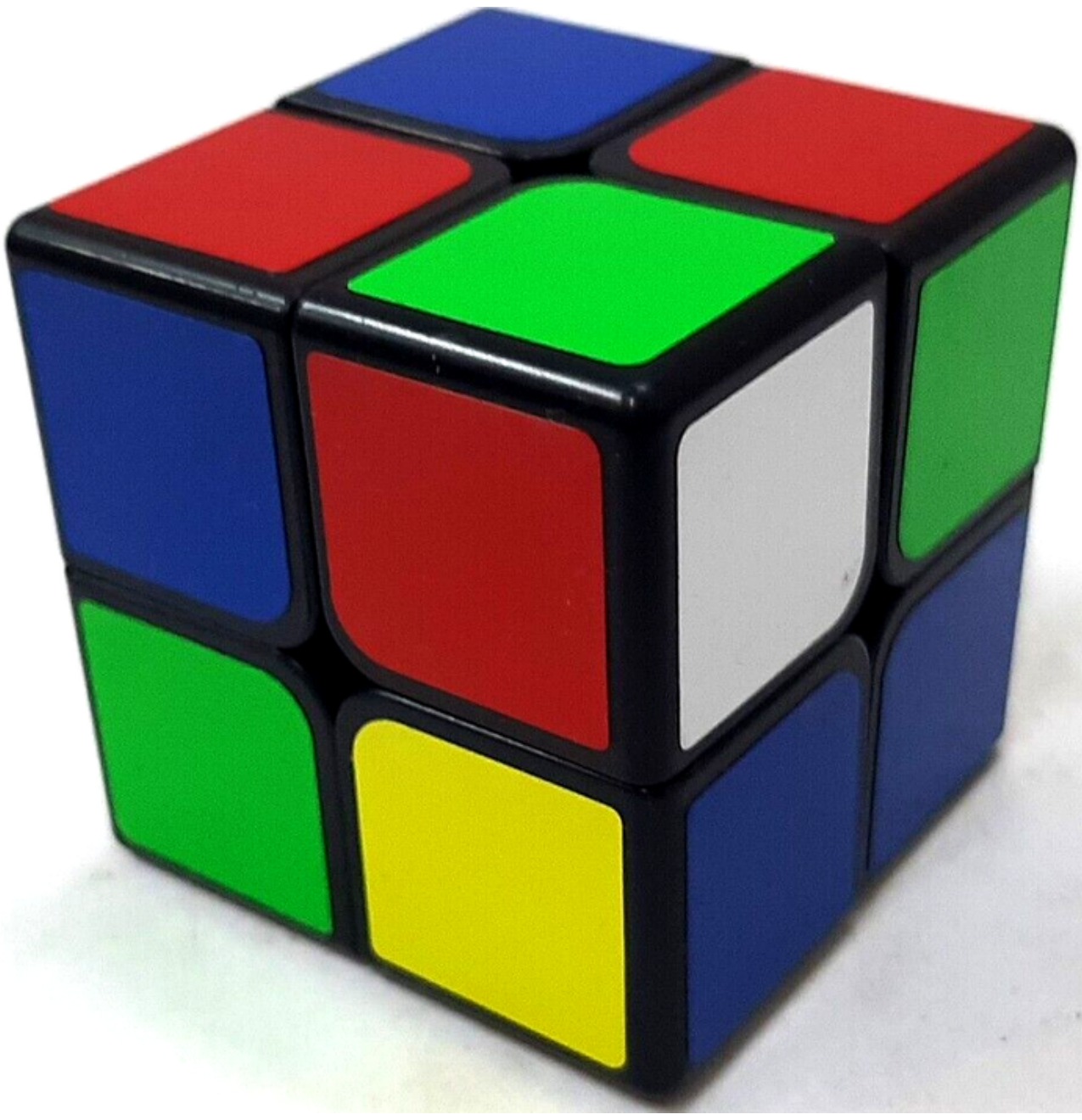}\end{tabular} \\ \midrule
State & \begin{tabular}[c]{@{}l@{}}The current configuration\\ of blocks\end{tabular} & \begin{tabular}[c]{@{}l@{}}The updated list of numbers \\ such as {[}1,2,3{]}\end{tabular} & \begin{tabular}[c]{@{}l@{}}Updated colors on \\ each face after a move\end{tabular} \\ \midrule
Action & \begin{tabular}[c]{@{}l@{}}Move, stack or unstack\\ blocks\end{tabular} & \begin{tabular}[c]{@{}l@{}}An arithmetic operation\\ between any two numbers\end{tabular} & \begin{tabular}[c]{@{}l@{}}One step rotation of \\ a Rubik's cube face\end{tabular} \\ \midrule
Output & A sequence of actions & \multicolumn{1}{l}{\begin{tabular}[l]{@{}l@{}}An equation evaluating to 24\\ for eg, $1\times2\times3\times4 = 24$\end{tabular}} & \begin{tabular}[c]{@{}l@{}}The sequence of rotations\\ to solve the Rubik's cube\end{tabular} \\ \bottomrule
\end{tabular}
\vspace{-0.2cm}
\end{table*}

\section{Experiments}

In this section, we evaluate AutoHD on three real-world planning tasks, including Blocksworld, Game of 24, and Rubik's Cube. The experimental results demonstrate that AutoHD significantly outperforms various baseline approaches. Additionally, we present an extensive ablation study in Section~\ref{subsec:ablation} to analyze the contributions of individual components. More experimental results can be found in Appendix~\ref{appendix:experimentsresults}.


\subsection{Blocksworld}
Blocksworld~\citep{valmeekam2022large} is a classic planning benchmark. It involves a set of blocks, each with unique identifiers, which can be moved or stacked according to specific rules. The goal is to transform an initial configuration of blocks into a specified target configuration using a series of predefined actions. Each action manipulates block positions while adhering to constraints such as only moving one block at a time and ensuring stability of stacks. States represent the current arrangement of blocks, while transitions occur as actions are applied, updating the state to reflect changes in block positions. LLMs need to demonstrate spatial reasoning capabilities to understand the physical interactions and constraints inherent in the task. 

We compare the proposed method with the following baseline methods, including (1) IO, which uses a standard input-output prompt; (2) CoT~\citep{wei2022chain}, a method that solves problems through a linear sequence of intermediate steps; (3) CoT-SC~\citep{wang2022self}, an extension of CoT that selects answers via majority voting across multiple paths; (4) ToT~\citep{yao2024tree}, which use tree search to explore and expand intermediate steps; (5) RAP~\citep{hao2023reasoning}, which integrates Monte Carlo Tree Search (MCTS) with the LLM as a world model, rewarding intermediate steps and guiding tree growth toward the correct answer; (6) LLM-Modulo~\citep{kambhampati2024llms}, which uses external critics, verifiers, and human input to ensure correctness of generated plans; and (7) AoT~\citep{sel2023algorithm}, which provides sequences of intermediate steps as in-context examples. Note that, we use 5 iterations for self-consistency. For O1 mini, we evaluate the model on a subset of Blockworld by randomly sampling 20 instances per step. 
We evaluate all methods using accuracy as the metric, which measures whether the action trace generated by LLMs successfully solves the task, i.e., rearranges the blocks to achieve the specified goal configuration. 

The results presented in Table~\ref{tab:bw} demonstrate that AutoHD consistently outperforms all baseline approaches across various LLMs, including GPT 4o-mini, GPT 4o, and Llama 3.1 70B, achieving accuracies of 42.4\%, 71.5\%, and 59.1\%, respectively. Notably, AutoHD achieves approximately twice the accuracy of the second-best baseline on GPT 4o and GPT 4o-mini. Additionally, while some baselines exhibit significant performance variance across different LLMs, AutoHD maintains robust and consistent results, highlighting the generalization ability of our approach. 
It is also worth noting that while O1 demonstrates near-perfect accuracy as an oracle model, AutoHD outperforms O1 mini, a state-of-the-art large reasoning model (LRM)~\citep{canLrmPlan}, further validating its effectiveness. 

\begin{table}[t]
    \centering
    \caption{Comparisons between AutoHD and baselines on Blocksworld. The best results are shown in bold.}
    \vspace{-0.2cm}
    \label{tab:bw}
    \begin{tabular}{l|cccc}
\toprule
\multirow{1}{*}{\centering Methods} 
 & \begin{tabular}[c]{@{}c@{}}GPT\\ 4o-mini\end{tabular} & \begin{tabular}[c]{@{}c@{}}GPT\\ 4o\end{tabular} & \begin{tabular}[c]{@{}c@{}}Llama 3.1\\ 70B\end{tabular} & \begin{tabular}[c]{@{}c@{}}O1\\ mini\end{tabular} \\ \hline
IO                       &   8.8      &    21.2          &      23.9        &  48.3 \\
CoT                      &    18.4        &       37.5       &     23.9      & - \\
CoT-SC@5                 &    21.2        &    41.5        &     30.7      & - \\
ToT                      &     23.1       &    12.4        &      4.6        & - \\
RAP      &   -        &    -       &     51.0      & - \\
LLM-Modulo & - & 48.0 & 13.0 & - \\
AoT     & - & 43.0 & 17.0 & - \\ 
AutoHD                     & \textbf{42.4} & \textbf{75.1} & \textbf{59.1} & - \\ \bottomrule
\end{tabular}
\vspace{-0.6cm}
\end{table}

\subsection{Game of 24}

The Game of 24 is a classic mathematical puzzle. It involves a set of four integers, typically drawn from a standard deck of playing cards, which can be combined using basic arithmetic operations, including addition, subtraction, multiplication, and division. The goal is to manipulate these numbers through a sequence of valid operations to produce the target value of $24$. Each state in the game represents the current set of intermediate results, while transitions occur as operations are applied, reducing the number of operands and updating the state. Constraints include the correct application of operations and adherence to mathematical rules, such as division by non-zero numbers. The Game of 24 is an NP-Complete problem and requires LLMs to use strong arithmetic reasoning to solve. 

We compare AutoHD with the following baseline methods, including IO, CoT~\citep{wei2022chain}, CoT-SC~\citep{wang2022self}, ToT~\citep{yao2024tree}, and AoT~\citep{sel2023algorithm}. Note that, we use 5 iterations for self-consistency.
We evaluate all methods using accuracy as the metric, which measures whether the action trace generated by LLMs successfully solves the task, i.e., forms the target number 24 by correctly applying a sequence of mathematical operations.

The results are shown in Table~\ref{tab:game24-results}. The proposed AutoHD outperforms all baselines consistently on three different LLM models. 
AutoHD achieves the highest accuracies of 54\%, 70\%, and 69\% for GPT 4o-mini, GPT 4o, and LLaMA 3.1 70B, respectively, significantly outperforming baselines. Simpler methods such as IO and CoT exhibit limited capability, achieving accuracies of at most 15\%. This highlights the challenges inherent in solving the Game of 24, which involves intricate numerical reasoning and arithmetic operations. Furthermore, AutoHD achieves results comparable to O1 mini, further demonstrating the strength and innovation of the proposed method.

\begin{table}[t]
\centering
\caption{Comparisons between AutoHD and baselines on Game of 24. The best results are shown in bold. }
\label{tab:game24-results}
\scalebox{1.0}{
\begin{tabular}{l|cccc}
\toprule
\multicolumn{1}{c|}{\multirow{1}{*}{Methods}}  & \begin{tabular}[c]{@{}c@{}}GPT \\ 4o-mini\end{tabular} & \begin{tabular}[c]{@{}c@{}}GPT\\ 4o\end{tabular} & \begin{tabular}[c]{@{}c@{}}LLaMA 3.1\\ 70B\end{tabular} & \begin{tabular}[c]{@{}c@{}}O1 \\ mini\end{tabular} \\ \midrule
IO & 9 & 8 & 3 & 77   \\
CoT & 13 & 14 & 8 & -  \\
CoT SC @ 5 & 15 & 14 & 8 & -  \\
ToT & 42 & 62 & 59 & -  \\
AutoHD & \textbf{54} & \textbf{70} & \textbf{69} & -  \\ \bottomrule
\end{tabular}
}
\end{table}

\begin{table}[t]
    \centering
    \caption{Comparisons between AutoHD and baselines on Rubik's Cube. The best results are shown in bold.}
    \label{tab:cube}
    \begin{tabular}{l|cccc}
\toprule
\multirow{1}{*}{Methods} 
 & \begin{tabular}[c]{@{}c@{}}GPT\\ 4o-mini\end{tabular} & \begin{tabular}[c]{@{}c@{}}GPT\\ 4o\end{tabular} & \begin{tabular}[c]{@{}c@{}}Llama 3.1\\ 70B\end{tabular} & \begin{tabular}[c]{@{}c@{}}O1\\ mini\end{tabular} \\ \hline
IO                       & 0.0           & 0.0             & 0.0             & 0.6 \\
CoT                      & 0.0           & 0.0             & 0.6          & - \\
CoT-SC@5                 & 0.0           & 0.6           & 0.6          & - \\
ToT                      &   0.6         &     9.8      &       13.1       & - \\
XoT                      & 67.2          & 79.8          & 78.1          & - \\
AutoHD                     & \textbf{82.5} & \textbf{83.1} & \textbf{84.7} & - \\ \bottomrule
\end{tabular}
\end{table}

\subsection{Rubik's Cube}

Rubik's cube~\citep{ding2023everything} is a well-known puzzle-solving benchmark requiring multi-step spatial planning. It involves a cube with six faces, each subdivided into four smaller squares of distinct colors. The goal is to transform an initial scrambled configuration into a target state where each face of the cube is uniformly colored. This is achieved through a sequence of predefined rotational actions applied to the cube’s faces. Each action corresponds to rotating one of the cube's layers along a specified axis, which alters the arrangement of colored squares adhering to the cube's structural constraints. States represent the current color arrangement of the cube, while transitions occur as rotations are executed, updating the state to reflect the new configuration.  Note that Rubik's cube is an NP-complete problem and the maximum number of steps required to optimally solve the cube is four in this dataset. 

We compare our AutoHD with the following baselines, including IO, CoT~\citep{wei2022chain}, CoT-SC~\citep{wang2022self}, ToT~\citep{yao2024tree}, and XoT~\citep{ding2023everything}. XoT uses reinforcement learning and MCTS to incorporate external domain knowledge by training an extra policy model on 1,000 training samples. 
We randomly select 15 examples from the training dataset in ~\citet{ding2023everything} to form the validation set used in the heuristic evolution process. We follow previous methods~\cite{ding2023everything,yu2024flow} to use a fixed state transition function instead of using LLM to update the states. We evaluate all methods using accuracy as the metric, which measures whether the action trace generated by LLMs successfully solves the task, i.e., transforms the cube into the goal state where each face is uniformly colored.


Results in Table~\ref{tab:cube} highlight the advance of AutoHD compared to baseline approaches on the Rubik's Cube task across various LLMs, including GPT 4o-mini, GPT 4o, Llama 3.1 70B. Specifically, AutoHD achieves the highest accuracies of 82.5\%, 83.1\%, and 84.7\% with GPT 4o-mini, GPT 4o, and Llama 3.1 70B, respectively, outperforming the strongest baseline, XoT, by substantial margins of 15.3\%, 3.3\%, and 6.6\%. Notably, simpler methods such as IO, CoT, and ToT struggle to solve the Rubik's Cube effectively, achieving nearly zero accuracy in most cases. This underscores the inherent difficulty of tasks like the Rubik's Cube, which require a deep understanding of complex spatial transformations. Even O1, a state-of-the-art large reasoning model, achieves only around 1\% accuracy, further emphasizing the challenges posed by the Rubik's Cube and the inability of existing LLMs to effectively capture spatial relationships.


\subsection{Ablation Studies}\label{subsec:ablation}
In this subsection, we conduct extensive ablation studies to analyze the contributions of different components. 

\textbf{Comparing Search Algorithms.} 
We begin by evaluating the performance of different search methods within our proposed framework. As discussed in Section~\ref{subsec:search}, in this work we study two search algorithms, including $\text{A}^*$ and greedy BFS. Experiments are conducted using GPT-4o mini on three tasks, namely Blocksworld, the Game of 24, and the Rubik's Cube. For Blocksworld, the dataset is divided into subsets based on the minimum number of actions required to solve each test case. Specifically, we evaluate different search methods on a subset corresponding to 2-step problems. The results, presented in Table~\ref{tab:searchAlg}, demonstrate that both $\text{A}^*$ and greedy BFS perform effectively within our framework, achieving comparable outcomes across different tasks.  These findings imply that LLM-generated heuristics are effective in guiding the search process during inference, highlighting the potential for future work to explore additional search algorithms.

\begin{table}[t]
\centering
\caption{Ablation study of different search algorithms using 4o-mini on Blocksworld, Game of 24, and Rubik's Cube.}
    \label{tab:searchAlg}
\begin{tabular}{l|ccccc}
 \toprule
 & \multicolumn{3}{c}{Blocksworld}                                  & \multirow{2}{*}{Game of 24} & \multirow{2}{*}{Rubik's cube} \\
 & 2 steps    & 4 steps & 6 steps                          &                               \\ \hline
BFS & \multicolumn{1}{c}{88.9} & 61.0 & 40.0 & 54.0  & 80.9 \\
$\text{A}^*$  & \multicolumn{1}{c}{95.7} & 73.3 & 69.2 & 50.0  & 82.5 \\
\bottomrule
\end{tabular}
\vspace{-0.2cm}
\end{table}


\textbf{Different LLMs for action generation and state transition.} We investigate the impact of using different LLMs for action generation and state transition on the performance of the proposed AutoHD. Specifically, we conduct experiments where heuristic functions are generated by a smaller model, GPT 4o-mini, while action generation and state transition are handled by a larger model, GPT-4o. In other words, the action space $\mathcal{A}_\theta$ and transition function $T_\theta$ are from GPT-4o. The results in Table~\ref{tab:worldmodel} show that AutoHD achieves significant improvements when using more powerful LLM for action generation and state transition. Moreover, our findings suggest that even smaller models, such as GPT-4o Mini, can generate effective heuristic functions. This observation implies that the primary limitation of LLMs in planning tasks may lie in their ability to generate actions and predict state transitions rather than in their capacity to produce heuristics.

\begin{table}[t]
\centering
\caption{Ablation study on different LLMs for action generation and state transition.}
    \label{tab:worldmodel}
\begin{tabular}{cc|cccc}
 \toprule
 \begin{tabular}[c]{@{}c@{}}Heuristic\\ generator\end{tabular}  & \begin{tabular}[c]{@{}c@{}}$\mathcal{A}_\theta\  \& \ T_\theta $  \\ \end{tabular} & \multicolumn{3}{c}{Blocksworld}                                  & \multirow{2}{*}{Game of 24}  \\
& & 2 steps & \multicolumn{1}{l}{4 steps} & \multicolumn{1}{l}{6 steps} &                                  \\ \hline
 4o-mini&  4o-mini & \multicolumn{1}{c}{95.7} & 73.3 & 69.2 &  54.0 \\
 4o-mini & 4o & \multicolumn{1}{c}{100.0} & 98.8 & 84.9 & 59.0 \\
 4o &  4o & \multicolumn{1}{c}{97.8} &96.4  & 92.8 & 70.0 \\
\bottomrule
\end{tabular}
\vspace{-0.5cm}
\end{table}

\begin{figure}
    \centering
    \includegraphics[width=0.6\linewidth]{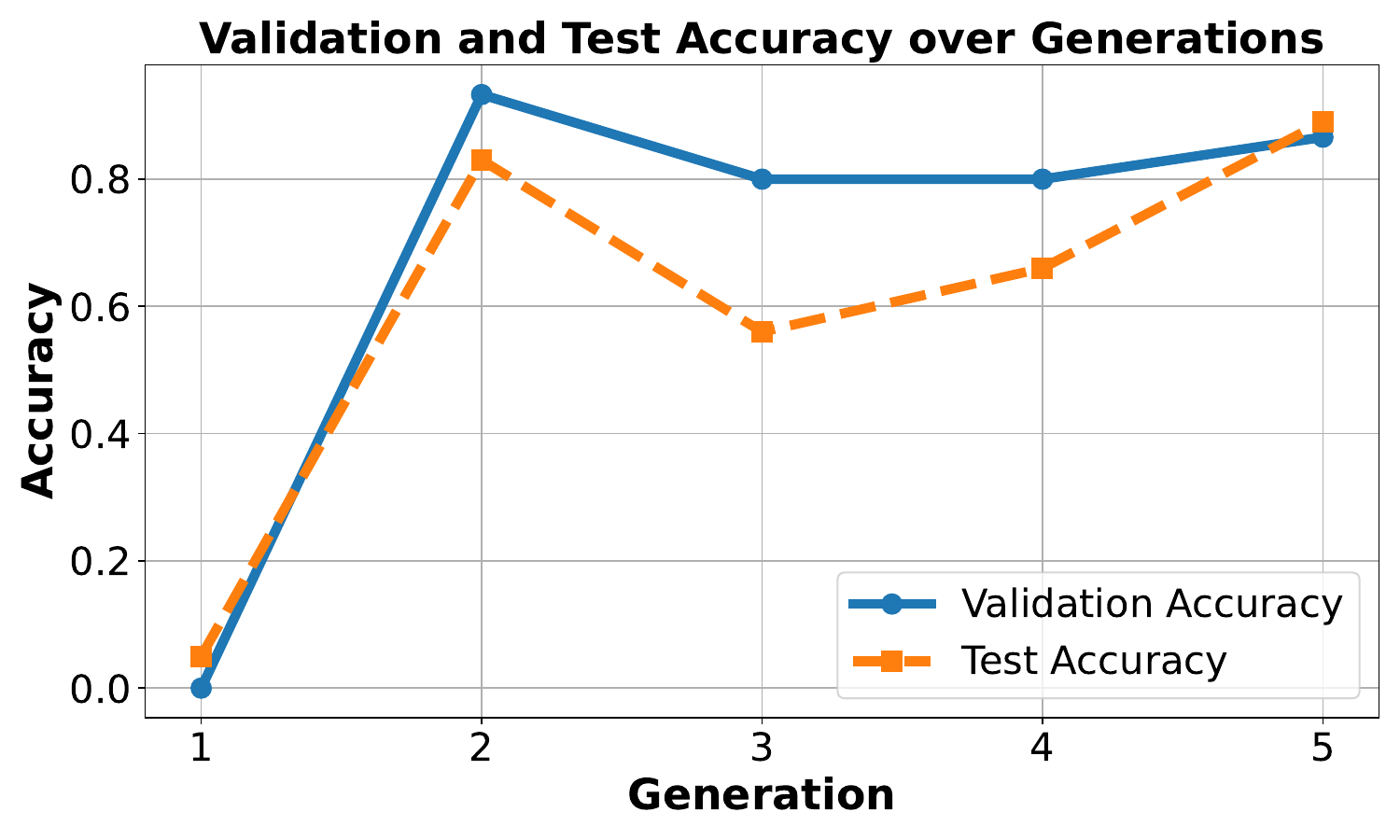}
    \vspace{-0.0cm}
    \caption{\small Ablation study of heuristic evolution on Rubik's Cube dataset.  The evolutionary process improves validation and test accuracies significantly in early generations before plateauing, indicating the discovery of a robust heuristic function.}
    \label{fig:evolution}
    \vspace{-0.0cm}
\end{figure}

\textbf{Heuristic Evolutions.} We also conduct experiments to analyze the evolution of heuristic functions. For each generation, we select the heuristic functions with the highest validation accuracy and evaluate their performance on the test dataset using GPT 4o-mini. The experiments are conducted on the Rubik's Cube dataset. 
The results in Figure~\ref{fig:evolution} show that both validation and test accuracies initially improve significantly during the early generations. Furthermore, performance eventually plateaus, indicating a saturation point in the evolution process. These suggest that the evolutionary process effectively explores the space of heuristic functions, ultimately generating a robust and well-performing heuristic function.
    
\begin{table}[t]
\centering
\caption{Results on multiple solutions on Game of 24.}
\label{tab:multi-solutions}
\scalebox{1.0}{
\begin{tabular}{l|ccc}
\toprule
\multicolumn{1}{c|}{\multirow{1}{*}{}} &  4o-mini &  4o & LLaMA 70B \\ \midrule
ToT & 47 & 68 & 61   \\ 
AutoHD & 66 & 92 & 86\\ \bottomrule
\end{tabular}
}
\vspace{-0.6cm}
\end{table}

\textbf{Multiple Solutions.} In complex planning tasks, multiple solutions often exist for a single problem. To assess how AutoHD handles such cases, we evaluate it on the Game of 24. For each test instance, we generate five solutions and consider it correct if at least one is valid. As shown in Table 7, AutoHD finds multiple correct solutions. Unlike ToT, which relies on LLM self-evaluation during the search, the LLM-generated heuristic in AutoHD helps the LLM efficiently find multiple solutions.

\section{Conclusions}

In this paper, we propose AutoHD, a novel framework that enables LLMs to explicitly generate heuristic functions for guiding inference-time search. Moreover, the heuristic evolution process further refines these functions, enhancing their robustness and effectiveness. The proposed AutoHD requires no additional model training or fine-tuning, making it adaptable to various tasks. The explicit heuristic functions generated by LLMs provide valuable insights into the reasoning process, making AutoHD a transparent solution for complex decision-making. Extensive experimentation across diverse benchmarks has validated the efficacy of our approach, showing substantial performance improvements over multiple baselines. These results firmly establish AutoHD as a reliable and interpretable solution for addressing complex planning tasks.

\section*{Acknowledgments}
This work was supported in part by National Science Foundation under grant CNS-2328395 and National Institutes of Health under grant U01AG070112.

\bibliography{llm}

\begin{thebibliography}{39}
\providecommand{\natexlab}[1]{#1}
\providecommand{\url}[1]{\texttt{#1}}
\expandafter\ifx\csname urlstyle\endcsname\relax
  \providecommand{\doi}[1]{doi: #1}\else
  \providecommand{\doi}{doi: \begingroup \urlstyle{rm}\Url}\fi

\bibitem[Besta et~al.(2024)Besta, Blach, Kubicek, Gerstenberger, Podstawski, Gianinazzi, Gajda, Lehmann, Niewiadomski, Nyczyk, et~al.]{besta2024graph}
Maciej Besta, Nils Blach, Ales Kubicek, Robert Gerstenberger, Michal Podstawski, Lukas Gianinazzi, Joanna Gajda, Tomasz Lehmann, Hubert Niewiadomski, Piotr Nyczyk, et~al.
\newblock Graph of thoughts: Solving elaborate problems with large language models.
\newblock In \emph{Proceedings of the AAAI Conference on Artificial Intelligence}, volume~38, pages 17682--17690, 2024.

\bibitem[Cobbe et~al.(2021)Cobbe, Kosaraju, Bavarian, Chen, Jun, Kaiser, Plappert, Tworek, Hilton, Nakano, Hesse, and Schulman]{cobbe2021gsm8k}
Karl Cobbe, Vineet Kosaraju, Mohammad Bavarian, Mark Chen, Heewoo Jun, Lukasz Kaiser, Matthias Plappert, Jerry Tworek, Jacob Hilton, Reiichiro Nakano, Christopher Hesse, and John Schulman.
\newblock Training verifiers to solve math word problems.
\newblock \emph{arXiv preprint arXiv:2110.14168}, 2021.

\bibitem[Ding et~al.(2023)Ding, Zhang, Wang, Xu, Ma, Zhang, Qin, Rajmohan, Lin, and Zhang]{ding2023everything}
Ruomeng Ding, Chaoyun Zhang, Lu~Wang, Yong Xu, Minghua Ma, Wei Zhang, Si~Qin, Saravan Rajmohan, Qingwei Lin, and Dongmei Zhang.
\newblock Everything of thoughts: Defying the law of penrose triangle for thought generation.
\newblock \emph{arXiv preprint arXiv:2311.04254}, 2023.

\bibitem[Driess et~al.(2023)Driess, Xia, Sajjadi, Lynch, Chowdhery, Ichter, Wahid, Tompson, Vuong, Yu, et~al.]{driess2023palm}
Danny Driess, Fei Xia, Mehdi~SM Sajjadi, Corey Lynch, Aakanksha Chowdhery, Brian Ichter, Ayzaan Wahid, Jonathan Tompson, Quan Vuong, Tianhe Yu, et~al.
\newblock Palm-e: An embodied multimodal language model.
\newblock \emph{arXiv preprint arXiv:2303.03378}, 2023.

\bibitem[Fan et~al.(2022)Fan, Wang, Jiang, Mandlekar, Yang, Zhu, Tang, Huang, Zhu, and Anandkumar]{fan2022minedojo}
Linxi Fan, Guanzhi Wang, Yunfan Jiang, Ajay Mandlekar, Yuncong Yang, Haoyi Zhu, Andrew Tang, De-An Huang, Yuke Zhu, and Anima Anandkumar.
\newblock Minedojo: Building open-ended embodied agents with internet-scale knowledge.
\newblock \emph{Advances in Neural Information Processing Systems}, 35:\penalty0 18343--18362, 2022.

\bibitem[Hao et~al.(2023)Hao, Gu, Ma, Hong, Wang, Wang, and Hu]{hao2023reasoning}
Shibo Hao, Yi~Gu, Haodi Ma, Joshua~Jiahua Hong, Zhen Wang, Daisy~Zhe Wang, and Zhiting Hu.
\newblock Reasoning with language model is planning with world model.
\newblock \emph{arXiv preprint arXiv:2305.14992}, 2023.

\bibitem[Hjeij and Vilks(2023)]{hjeij2023brief}
Mohamad Hjeij and Arnis Vilks.
\newblock A brief history of heuristics: how did research on heuristics evolve?
\newblock \emph{Humanities and Social Sciences Communications}, 10\penalty0 (1):\penalty0 1--15, 2023.

\bibitem[Huang et~al.(2023)Huang, Chen, Mishra, Zheng, Yu, Song, and Zhou]{huang2023large}
Jie Huang, Xinyun Chen, Swaroop Mishra, Huaixiu~Steven Zheng, Adams~Wei Yu, Xinying Song, and Denny Zhou.
\newblock Large language models cannot self-correct reasoning yet.
\newblock \emph{arXiv preprint arXiv:2310.01798}, 2023.

\bibitem[Huang et~al.(2024)Huang, Chen, Mishra, Zheng, Yu, Song, and Zhou]{huang2024large}
Jie Huang, Xinyun Chen, Swaroop Mishra, Huaixiu~Steven Zheng, Adams~Wei Yu, Xinying Song, and Denny Zhou.
\newblock Large language models cannot self-correct reasoning yet.
\newblock In \emph{The Twelfth International Conference on Learning Representations}, 2024.
\newblock URL \url{https://openreview.net/forum?id=IkmD3fKBPQ}.

\bibitem[Huang et~al.(2022)Huang, Xia, Xiao, Chan, Liang, Florence, Zeng, Tompson, Mordatch, Chebotar, et~al.]{huang2022inner}
Wenlong Huang, Fei Xia, Ted Xiao, Harris Chan, Jacky Liang, Pete Florence, Andy Zeng, Jonathan Tompson, Igor Mordatch, Yevgen Chebotar, et~al.
\newblock Inner monologue: Embodied reasoning through planning with language models.
\newblock \emph{arXiv preprint arXiv:2207.05608}, 2022.

\bibitem[Kambhampati et~al.(2024)Kambhampati, Valmeekam, Guan, Verma, Stechly, Bhambri, Saldyt, and Murthy]{kambhampati2024llms}
Subbarao Kambhampati, Karthik Valmeekam, Lin Guan, Mudit Verma, Kaya Stechly, Siddhant Bhambri, Lucas Saldyt, and Anil Murthy.
\newblock Llms can't plan, but can help planning in llm-modulo frameworks.
\newblock \emph{arXiv preprint arXiv:2402.01817}, 2024.

\bibitem[Kannan et~al.(2024)Kannan, Venkatesh, and Min]{kannan2024smart}
Shyam~Sundar Kannan, Vishnunandan~LN Venkatesh, and Byung-Cheol Min.
\newblock Smart-llm: Smart multi-agent robot task planning using large language models.
\newblock In \emph{2024 IEEE/RSJ International Conference on Intelligent Robots and Systems (IROS)}, pages 12140--12147. IEEE, 2024.

\bibitem[Koza(1994)]{koza1994genetic}
John~R Koza.
\newblock Genetic programming as a means for programming computers by natural selection.
\newblock \emph{Statistics and computing}, 4:\penalty0 87--112, 1994.

\bibitem[Li et~al.(2022)Li, Lin, Zhang, Fu, Chen, Lou, and Chen]{li2022making}
Yifei Li, Zeqi Lin, Shizhuo Zhang, Qiang Fu, Bei Chen, Jian-Guang Lou, and Weizhu Chen.
\newblock Making large language models better reasoners with step-aware verifier.
\newblock \emph{arXiv preprint arXiv:2206.02336}, 2022.

\bibitem[Lightman et~al.(2023)Lightman, Kosaraju, Burda, Edwards, Baker, Lee, Leike, Schulman, Sutskever, and Cobbe]{lightman2023let}
Hunter Lightman, Vineet Kosaraju, Yura Burda, Harri Edwards, Bowen Baker, Teddy Lee, Jan Leike, John Schulman, Ilya Sutskever, and Karl Cobbe.
\newblock Let's verify step by step.
\newblock \emph{arXiv preprint arXiv:2305.20050}, 2023.

\bibitem[Liu et~al.(2024{\natexlab{a}})Liu, Xialiang, Yuan, Lin, Luo, Wang, Lu, and Zhang]{liu2024evolution}
Fei Liu, Tong Xialiang, Mingxuan Yuan, Xi~Lin, Fu~Luo, Zhenkun Wang, Zhichao Lu, and Qingfu Zhang.
\newblock Evolution of heuristics: Towards efficient automatic algorithm design using large language model.
\newblock In \emph{Forty-first International Conference on Machine Learning}, 2024{\natexlab{a}}.

\bibitem[Liu et~al.(2024{\natexlab{b}})Liu, Chen, Qu, Tang, and Ong]{liu2024large}
Shengcai Liu, Caishun Chen, Xinghua Qu, Ke~Tang, and Yew-Soon Ong.
\newblock Large language models as evolutionary optimizers.
\newblock In \emph{2024 IEEE Congress on Evolutionary Computation (CEC)}, pages 1--8. IEEE, 2024{\natexlab{b}}.

\bibitem[Liu et~al.(2023)Liu, Yu, Zhang, Xu, Lei, Lai, Gu, Ding, Men, Yang, et~al.]{liu2023agentbench}
Xiao Liu, Hao Yu, Hanchen Zhang, Yifan Xu, Xuanyu Lei, Hanyu Lai, Yu~Gu, Hangliang Ding, Kaiwen Men, Kejuan Yang, et~al.
\newblock Agentbench: Evaluating llms as agents.
\newblock \emph{arXiv preprint arXiv:2308.03688}, 2023.

\bibitem[Luo et~al.(2024)Luo, Liu, Liu, Phatale, Lara, Li, Shu, Zhu, Meng, Sun, et~al.]{luo2024improve}
Liangchen Luo, Yinxiao Liu, Rosanne Liu, Samrat Phatale, Harsh Lara, Yunxuan Li, Lei Shu, Yun Zhu, Lei Meng, Jiao Sun, et~al.
\newblock Improve mathematical reasoning in language models by automated process supervision.
\newblock \emph{arXiv preprint arXiv:2406.06592}, 2024.

\bibitem[Parashar et~al.(2024)Parashar, Lin, Liu, Dong, Li, Ramanan, Caverlee, and Kong]{parashar2024neglected}
Shubham Parashar, Zhiqiu Lin, Tian Liu, Xiangjue Dong, Yanan Li, Deva Ramanan, James Caverlee, and Shu Kong.
\newblock The neglected tails in vision-language models.
\newblock In \emph{Proceedings of the IEEE/CVF Conference on Computer Vision and Pattern Recognition}, pages 12988--12997, 2024.

\bibitem[Parashar et~al.(2025)Parashar, Olson, Khurana, Li, Ling, Caverlee, and Ji]{2025Sys2BenchLLM}
Shubham Parashar, Blake Olson, Sambhav Khurana, Eric Li, Hongyi Ling, James Caverlee, and Shuiwang Ji.
\newblock Inference-time computations for llm reasoning and planning: A benchmark and insights, 2025.
\newblock URL \url{https://arxiv.org/abs/2502.12521}.

\bibitem[Romera-Paredes et~al.(2024)Romera-Paredes, Barekatain, Novikov, Balog, Kumar, Dupont, Ruiz, Ellenberg, Wang, Fawzi, et~al.]{romera2024mathematical}
Bernardino Romera-Paredes, Mohammadamin Barekatain, Alexander Novikov, Matej Balog, M~Pawan Kumar, Emilien Dupont, Francisco~JR Ruiz, Jordan~S Ellenberg, Pengming Wang, Omar Fawzi, et~al.
\newblock Mathematical discoveries from program search with large language models.
\newblock \emph{Nature}, 625\penalty0 (7995):\penalty0 468--475, 2024.

\bibitem[Saparov and He(2023)]{PrOntoQA}
Abulhair Saparov and He~He.
\newblock Language models are greedy reasoners: A systematic formal analysis of chain-of-thought.
\newblock In \emph{The Eleventh International Conference on Learning Representations}, 2023.
\newblock URL \url{https://openreview.net/forum?id=qFVVBzXxR2V}.

\bibitem[Sel et~al.(2023)Sel, Al-Tawaha, Khattar, Jia, and Jin]{sel2023algorithm}
Bilgehan Sel, Ahmad Al-Tawaha, Vanshaj Khattar, Ruoxi Jia, and Ming Jin.
\newblock Algorithm of thoughts: Enhancing exploration of ideas in large language models.
\newblock \emph{arXiv preprint arXiv:2308.10379}, 2023.

\bibitem[Shinn et~al.(2024)Shinn, Cassano, Gopinath, Narasimhan, and Yao]{shinn2024reflexion}
Noah Shinn, Federico Cassano, Ashwin Gopinath, Karthik Narasimhan, and Shunyu Yao.
\newblock Reflexion: Language agents with verbal reinforcement learning.
\newblock \emph{Advances in Neural Information Processing Systems}, 36, 2024.

\bibitem[Simon(1997)]{simon1997models}
Herbert~A Simon.
\newblock \emph{Models of Bounded Rationality: Empirically Grounded Economic Reason}.
\newblock The MIT Press, 1997.

\bibitem[Singh et~al.(2023)Singh, Blukis, Mousavian, Goyal, Xu, Tremblay, Fox, Thomason, and Garg]{singh2023progprompt}
Ishika Singh, Valts Blukis, Arsalan Mousavian, Ankit Goyal, Danfei Xu, Jonathan Tremblay, Dieter Fox, Jesse Thomason, and Animesh Garg.
\newblock Progprompt: Generating situated robot task plans using large language models.
\newblock In \emph{2023 IEEE International Conference on Robotics and Automation (ICRA)}, pages 11523--11530. IEEE, 2023.

\bibitem[Stechly et~al.(2024)Stechly, Valmeekam, and Kambhampati]{stechly2024self}
Kaya Stechly, Karthik Valmeekam, and Subbarao Kambhampati.
\newblock On the self-verification limitations of large language models on reasoning and planning tasks.
\newblock \emph{arXiv preprint arXiv:2402.08115}, 2024.

\bibitem[Tversky and Kahneman(1974)]{tversky1974judgment}
Amos Tversky and Daniel Kahneman.
\newblock Judgment under uncertainty: Heuristics and biases: Biases in judgments reveal some heuristics of thinking under uncertainty.
\newblock \emph{science}, 185\penalty0 (4157):\penalty0 1124--1131, 1974.

\bibitem[Valmeekam et~al.(2022)Valmeekam, Olmo, Sreedharan, and Kambhampati]{valmeekam2022large}
Karthik Valmeekam, Alberto Olmo, Sarath Sreedharan, and Subbarao Kambhampati.
\newblock Large language models still can't plan (a benchmark for llms on planning and reasoning about change).
\newblock In \emph{NeurIPS 2022 Foundation Models for Decision Making Workshop}, 2022.

\bibitem[Valmeekam et~al.(2024)Valmeekam, Stechly, and Kambhampati]{canLrmPlan}
Karthik Valmeekam, Kaya Stechly, and Subbarao Kambhampati.
\newblock Llms still can't plan; can lrms? a preliminary evaluation of openai's o1 on planbench.
\newblock \emph{arXiv preprint arXiv:2409.13373}, 2024.

\bibitem[Veli{\v{c}}kovi{\'c} et~al.(2024)Veli{\v{c}}kovi{\'c}, Vitvitskyi, Markeeva, Ibarz, Buesing, Balog, and Novikov]{velivckovic2024amplifying}
Petar Veli{\v{c}}kovi{\'c}, Alex Vitvitskyi, Larisa Markeeva, Borja Ibarz, Lars Buesing, Matej Balog, and Alexander Novikov.
\newblock Amplifying human performance in combinatorial competitive programming.
\newblock \emph{arXiv preprint arXiv:2411.19744}, 2024.

\bibitem[Wang et~al.(2023)Wang, Xie, Jiang, Mandlekar, Xiao, Zhu, Fan, and Anandkumar]{wang2023voyager}
Guanzhi Wang, Yuqi Xie, Yunfan Jiang, Ajay Mandlekar, Chaowei Xiao, Yuke Zhu, Linxi Fan, and Anima Anandkumar.
\newblock Voyager: An open-ended embodied agent with large language models.
\newblock \emph{arXiv preprint arXiv:2305.16291}, 2023.

\bibitem[Wang et~al.(2022)Wang, Wei, Schuurmans, Le, Chi, Narang, Chowdhery, and Zhou]{wang2022self}
Xuezhi Wang, Jason Wei, Dale Schuurmans, Quoc Le, Ed~Chi, Sharan Narang, Aakanksha Chowdhery, and Denny Zhou.
\newblock Self-consistency improves chain of thought reasoning in language models.
\newblock \emph{arXiv preprint arXiv:2203.11171}, 2022.

\bibitem[Wei et~al.(2022)Wei, Wang, Schuurmans, Bosma, Xia, Chi, Le, Zhou, et~al.]{wei2022chain}
Jason Wei, Xuezhi Wang, Dale Schuurmans, Maarten Bosma, Fei Xia, Ed~Chi, Quoc~V Le, Denny Zhou, et~al.
\newblock Chain-of-thought prompting elicits reasoning in large language models.
\newblock \emph{Advances in neural information processing systems}, 35:\penalty0 24824--24837, 2022.

\bibitem[Yang et~al.(2018)Yang, Qi, Zhang, Bengio, Cohen, Salakhutdinov, and Manning]{yang2018hotpotqa}
Zhilin Yang, Peng Qi, Saizheng Zhang, Yoshua Bengio, William~W Cohen, Ruslan Salakhutdinov, and Christopher~D Manning.
\newblock Hotpotqa: A dataset for diverse, explainable multi-hop question answering.
\newblock \emph{arXiv preprint arXiv:1809.09600}, 2018.

\bibitem[Yao et~al.(2022)Yao, Zhao, Yu, Du, Shafran, Narasimhan, and Cao]{yao2022react}
Shunyu Yao, Jeffrey Zhao, Dian Yu, Nan Du, Izhak Shafran, Karthik Narasimhan, and Yuan Cao.
\newblock React: Synergizing reasoning and acting in language models.
\newblock \emph{arXiv preprint arXiv:2210.03629}, 2022.

\bibitem[Yao et~al.(2024)Yao, Yu, Zhao, Shafran, Griffiths, Cao, and Narasimhan]{yao2024tree}
Shunyu Yao, Dian Yu, Jeffrey Zhao, Izhak Shafran, Tom Griffiths, Yuan Cao, and Karthik Narasimhan.
\newblock Tree of thoughts: Deliberate problem solving with large language models.
\newblock \emph{Advances in Neural Information Processing Systems}, 36, 2024.

\bibitem[Yu et~al.(2024)Yu, Jiang, Kang, Hao, and Qin]{yu2024flow}
Fangxu Yu, Lai Jiang, Haoqiang Kang, Shibo Hao, and Lianhui Qin.
\newblock Flow of reasoning: Efficient training of llm policy with divergent thinking.
\newblock \emph{arXiv preprint arXiv:2406.05673}, 2024.

\end{thebibliography}
\bibliographystyle{plainnat}

\newpage
\appendix
\section{Prompts}
In this section, we provide the detailed prompts used in our experiments.

\subsection{Prompts for Heuristic Function Proposal}\label{appendix:heuristic_proposal_prompt}
In this subsection, we show the prompts we used to get the initial heuristics For Blocksworld, Game of 24, and Rubik's Cube. 

\begin{tcolorbox}[
    colback=black!10!white, 
    colframe=black,         
    rounded corners,        
    title=Prompt for Heuristic Function Proposal for Blocksworld,
]
I need help designing a new heuristic function to solve blocksworld problem. You are given 2 strings - current state of blocks and a final desired state. The goal is to design an appropriate heuristic that can be used to solve how far current state is from final state. \\

Here is an example problem:\\
Initial state: the red block is clear, the yellow block is clear, the hand is empty, the red block is on top of the blue block, the yellow block is on top of the orange block, the blue block is on the table and the orange block is on the table\\
Final state: the orange block is clear, the yellow block is clear, the hand is empty, the orange block is on top of the red block, the red block is on top of the blue block, the blue block is on the table, and the yellow block is on the table.\\

Here is another example problem:\\
Initial state: the blue block is clear, the orange block is in the hand, the red block is clear, the yellow block is clear, the hand is holding the orange block, the blue block is on the table, the red block is on the table, and the yellow block is on the table.\\
Final state: the orange block is clear, the red block is clear, the yellow block is clear, the hand is empty, the red block is on top of the blue block, the blue block is on the table, the orange block is on the table, and the yellow block is on the table.\\

Task:\\
Please design a new heuristic. \\
Firstly, describe your heuristic and main steps in one sentence. Start the sentence with `Heuristic Description:'\\

Next, implement it in Python as a function named `calc\_heuristic'. This function should accept 2 string inputs as shown above (comma separated with and at the last):\\
1. `initial\_state' - The current state of blocks.\\
2. `final\_state' - The final state of blocks.\\

This function should return one output: `heuristic\_val', which is the heuristic value calculated for the current state of the blocks with respect to final goal state.\\

Do not give additional explanations.
\end{tcolorbox}
\begin{tcolorbox}[
    colback=black!10!white, 
    colframe=black,         
    rounded corners,        
    title=Prompt for Heuristic Function Proposal for Game of 24,
]
I need help designing a new heuristic function to solve Game 24 problem. In Game 24, you are given a list of numbers that need to be used with any operation '+', '-', '*', '/' to obtain the goal, which is [24]. You need to design an appropriate heuristic that can be used to solve how far current state is from final state.  \\

Here is an example problem:
Current State: [4, 4, 6, 8] \ \# (4 + 8) * (6 - 4) = 24 \\
Final State: [24]\\

Here is another example problem:
Current State: [4, 6] \# 4 * 6 = 24 \ \\
Final State: [24]\\

Here is another example problem:
Current State: [8, 4, 1, 8] \ \# (8 / 4 + 1) * 8 = 24 \\
Final State: [24]\\

Here is another example problem:
Current State: [5, 5, 5, 9] \ \# 5 + 5 + 5 + 9 = 24 \\
Final State: [24]\\

Here is another example problem:
Current State: [24] \\
Final State: [24]\\

Task:\\
Please design a new heuristic. \\
Firstly, describe your heuristic and main steps in one sentence as a python comment. Start the comment with `Heuristic Description:'\\

Next, implement it in Python as a function named `calc\_heuristic'. This function should accept 1 argument as show below, and not modify the input:\\
1. 'Numbers' - The current state, a list of numbers that has to be used to obtain the goal. \\

This function should return a single output, heuristic\_val, representing the feasibility of reaching the goal, considering any operation using (+, -, *, /) with the current state. For eg: (a+b, a-b, b-a, a*b, a/b, b/a). \\
It must return 0 if the goal is achieved, i.e. when 24 is the only number left. \\



Do not give additional explanations. Do not use any tools. Return your response as python code.
\end{tcolorbox}

\begin{tcolorbox}[
    colback=black!10!white, 
    colframe=black,         
    rounded corners,        
    title=Prompt for Heuristic Function Proposal for Rubik's Cube,
]
I need help designing a new heuristic function to solve 2x2 Pocket Cube. The problem is defined as the following. Your task is to restore a scrambled 2x2 Rubik's Cube to its original state. All the given problems can be solved in 1 to 4 moves. You cannot exceed more than 11 moves. Provide the sequence of moves required for the restoration. Please follow the instructions and rules below to complete the solving:\\
1. A 2x2 Pocket Cube has six faces, namely: [Upper, Front, Bottom, Left, Right, Back] Each consisting of a 2x2 grid of squares, with each square having its own color.\\
2. Colors in the Cube are represented in numbers: [0, 1, 2, 3, 4, 5]\\
3. The Cube's state is represented as an array of 24 elements. For instance, [0,0,0,0,1,1,1,1,2,2,2,2,3,3,3,3,4,4,4,4,5,5,5,5]. The Cube's state is represented as a 24-element array, where each group of 4 consecutive elements corresponds to a face of the cube in the following order: Upper face: Elements at indices 0 to 3. Right face: Elements at indices 4 to 7. Front face: Elements at indices 8 to 11. Down face: Elements at indices 12 to 15. Left face: Elements at indices 16 to 19. Back face: Elements at indices 20 to 23. Each element within a group represents the color or state of a specific square on that face.\\
4. A restoration of a Pocket Cube is to move squares in each face to have same numbers. Some example Restored States are [0,0,0,0,1,1,1,1,2,2,2,2,3,3,3,3,4,4,4,4,5,5,5,5]. \\
You must make move to the Cube to achieve a Restored State, not limited to the above one. Note that we just need each face to have same numbers, no matter which face has which color.\\
Task:\\
Please design a new heuristic. \\
Firstly, describe your heuristic and main steps in one sentence. Start the sentence with `Heuristic Description:'\\

Next, implement it in Python as a function named `calc\_heuristic'. This function should accept 1 input as shown above :\\
1. `State' - The current state of 2x2 Cube, which is a numpy array.\\

This function should return one output: `heuristic\_val', which is the heuristic value calculated for the current state of the 2x2 Cube with respect to one of restored states.\\

Do not give additional explanations. 
\end{tcolorbox}

\subsection{Prompts for Heuristic Evolution}\label{appendix:heuristic_evolution_prompt}

\begin{tcolorbox}[
    colback=black!10!white, 
    colframe=black,         
    rounded corners,        
    title=Prompt for Heuristic Evolution for Blocksworld,
]
I need help designing a new heuristic function to solve blocksworld problem. You are given 2 strings - current state of blocks and a final desired state. The goal is to design an appropriate heuristic that can be used to solve how far current state is from final state. \\

Here is an example problem:\\
Initial state: the red block is clear, the yellow block is clear, the hand is empty, the red block is on top of the blue block, the yellow block is on top of the orange block, the blue block is on the table and the orange block is on the table\\
Final state: the orange block is clear, the yellow block is clear, the hand is empty, the orange block is on top of the red block, the red block is on top of the blue block, the blue block is on the table, and the yellow block is on the table.\\

Here is another example problem:\\
Initial state: the blue block is clear, the orange block is in the hand, the red block is clear, the yellow block is clear, the hand is holding the orange block, the blue block is on the table, the red block is on the table, and the yellow block is on the table.\\
Final state: the orange block is clear, the red block is clear, the yellow block is clear, the hand is empty, the red block is on top of the blue block, the blue block is on the table, the orange block is on the table, and the yellow block is on the table.\\

$<$exisiting\_heuristics$>$\\

Task:\\
$<$evolution\_type$>$\\
Firstly, identify the common idea in the provided heuristics.\\
Secondly, based on the backbone idea describe your new heuristic in one sentence.\\
Thirdly, implement it in Python as a function named `calc\_heuristic'. This function should accept 2 string inputs as shown above (comma separated with and at the last):\\
1. `initial\_state' - The current state of blocks.\\
2. `final\_state' - The final state of blocks.\\

This function should return one output: `heuristic\_val', which is the heuristic value calculated for the current state of the blocks.\\

Do not give additional explanations. 
\end{tcolorbox}

\begin{tcolorbox}[
    colback=black!10!white, 
    colframe=black,         
    rounded corners,        
    title=Prompt for Heuristic Evolution for Game of 24,
]
I need help designing a new heuristic function to solve Game 24 problem. In Game 24, you are given a list of numbers that need to be used with any operation '+', '-', '*', '/' to obtain the goal, which is [24]. You need to design an appropriate heuristic that can be used to solve how far current state is from final state. \\

Here is an example problem:
Current State: [4, 4, 6, 8] \ \# (4 + 8) * (6 - 4) = 24 \\
Final State: [24] \\

Here is another example problem:
Current State: [4, 6] \ \# 4 * 6 = 24 \\ 
Final State: [24] \\

Here is another example problem:
Current State: [8, 4, 1, 8] \ \# (8 / 4 + 1) * 8 = 24 \\
Final State: [24] \\

Here is another example problem:
Current State: [5, 5, 5, 9] \ \# 5 + 5 + 5 + 9 = 24 \\
Final State: [24] \\

Here is another example problem:
Current State: [24] \\
Final State: [24] \\

$<$exisiting\_heuristics$>$ \\

Task: \\
$<$evolution\_type$>$ \\

Firstly, identify the common idea in the provided heuristics in one sentence as python comment. Start the python comment with 'Common Idea:' \\
Secondly, based on the backbone idea describe your new heuristic in one sentence as a python comment. Start the python comment with 'Heuristic Description:' \\
Thirdly, implement it in Python as a function named 'calc\_heuristic'. This function should accept 1 argument as show below: \\
1. 'Numbers' - The current state, a list of numbers that has to be used to obtain the goal. \\

This function should return a single output, heuristic\_val, representing the feasibility of reaching the goal, considering any operation using (+, -, *, /) with the current state. For eg: (a+b, a-b, b-a, a*b, a/b, b/a). \\
It must return 0 if the goal is achieved, i.e. when 24 is the only number left. \\

Do not give additional explanations. Do not use any tools. Return your response as python code.\\
\end{tcolorbox}

\begin{tcolorbox}[
    colback=black!10!white, 
    colframe=black,         
    rounded corners,        
    title=Prompt for Heuristic Evolution for Rubik's Cube,
]
I need help designing a new heuristic function to solve 2x2 Pocket Cube. The problem is defined as the following. Your task is to restore a scrambled 2x2 Rubik's Cube to its original state. All the given problems can be solved in 1 to 4 moves. You cannot exceed more than 11 moves. Provide the sequence of moves required for the restoration. Please follow the instructions and rules below to complete the solving:\\
1. A 2x2 Pocket Cube has six faces, namely: [Upper, Front, Bottom, Left, Right, Back] Each consisting of a 2x2 grid of squares, with each square having its own color.\\
2. Colors in the Cube are represented in numbers: [0, 1, 2, 3, 4, 5]\\
3. The Cube's state is represented as an array of 24 elements. For instance, [0,0,0,0,1,1,1,1,2,2,2,2,3,3,3,3,4,4,4,4,5,5,5,5]. The Cube's state is represented as a 24-element array, where each group of 4 consecutive elements corresponds to a face of the cube in the following order: Upper face: Elements at indices 0 to 3. Right face: Elements at indices 4 to 7. Front face: Elements at indices 8 to 11. Down face: Elements at indices 12 to 15. Left face: Elements at indices 16 to 19. Back face: Elements at indices 20 to 23. Each element within a group represents the color or state of a specific square on that face.\\
4. A restoration of a Pocket Cube is to move squares in each face to have same numbers. Some example Restored States are [0,0,0,0,1,1,1,1,2,2,2,2,3,3,3,3,4,4,4,4,5,5,5,5]. \\
You must make move to the Cube to achieve a Restored State, not limited to the above one. Note that we just need each face to have same numbers, no matter which face has which color.\\

$<$exisiting\_heuristics$>$\\

Task:\\
$<$evolution\_type$>$\\
Firstly, identify the common idea in the provided heuristics.\\
Secondly, based on the backbone idea describe your new heuristic in one sentence.\\
Thirdly, implement it in Python as a function named `calc\_heuristic'. This function should accept 1 input as shown above :\\
1. `State' - The current state of 2x2 Cube, which is a numpy array.\\

This function should return one output: `heuristic\_val', which is the heuristic value calculated for the current state of the 2x2 Cube with respect to one of restored states.\\

Do not give additional explanations. 

\end{tcolorbox}

For the evolution type, we use following four prompts. 

\begin{tcolorbox}[
    colback=black!10!white, 
    colframe=black,         
    rounded corners,        
    title=Evolution Types,
]
1. Please help me create a new heuristic that has a totally different form from the given ones. \\
2. Please help me create a new heuristic that has a totally different form from the given ones but can be motivated from them. \\
3. Please assist me in creating a new heuristic that has a different form but can be a modified version of the heuristic provided. \\
4. Please identify the main heuristic parameters and assist me in creating a new heuristic that has a different parameter settings of the score function provided. \\

\end{tcolorbox}

\section{More Experimental Results}\label{appendix:experimentsresults}
In this section, we present additional experimental results. For the Blocksworld dataset, we divide the data into subsets based on the minimum number of actions required to solve each test case. The results are summarized in Table~\ref{tab:bw_split}.  
\begin{table}[t]
\centering
\caption{Comparisons between AutoHD and baselines on Blocksworld. The best results are shown in bold.}
    \label{tab:bw_split}
\begin{tabular}{c|l|ccccccc}
\toprule
& Methods & Step 2 & Step 4 & Step 6 & Step 8 & Step 10 & Step 12 & All \\ \hline
\multirow{5}{*}{\begin{tabular}[c]{@{}c@{}}GPT\\ 4o-mini\end{tabular}} & IO  &  46.7   &  25.0     & 5.9 & 0.7 & 0.0 & 0.0 & 8.8 \\
& CoT    & 38.3 & 26.7 & 26.5 & 13.1 & 8.0 & 0.0 & 18.4\\
& CoT-SC@5 & 55.3 & 36.1 & 30.3 & 12.4 & 4.4 & 0.0 & 21.2\\
& ToT    & 71.1 & 39.3 & 26.3 & 16.6 & \textbf{20.5} & \textbf{17.4} & 23.1\\
& AutoHD   & \textbf{95.7} & \textbf{73.3} & \textbf{69.2} & \textbf{20.9} & 8.0 & 0.0 & \textbf{42.4} \\ \hline
\multirow{5}{*}{\begin{tabular}[c]{@{}c@{}}GPT\\ 4o\end{tabular}} & IO & 51.1 & 39.3& 30.3 & 13.9 & 0.9 & 2.2 & 21.2\\
& CoT    & 68.1 & 45.4& 52.3 & 28.1 & 19.5 & 19.6 & 37.5\\
& CoT-SC@5 & 63.8 & 39.5& 57.4 & 35.3 & 26.5 & \textbf{28.2} & 41.5\\
& ToT    & 75.5 & 21.4& 10 & 1.3 & 2.7 & 2.2 & 12.4\\
& AutoHD   & \textbf{97.8} & \textbf{96.4} & \textbf{92.8} & \textbf{80.8} & \textbf{42.9} & 15.2& \textbf{75.1}\\ \hline
\multirow{5}{*}{\begin{tabular}[c]{@{}c@{}}Llama 3.1\\ 70B\end{tabular} } & IO  & 44.4 &  32.1 & 30.9 & 19.9 & 13.4 & 4.4 & 23.9\\
& CoT    & 46.7 & 44.1 & 28.3 & 19.9 & \textbf{18.8} &4.4 & 26.1\\
& CoT-SC@5 & 42.2& 47.0& 39.5 & 22.5 & 17.4 & \textbf{19.6} & 30.7\\
& ToT    & 60.0& 0.0& 0.0 & 0.0 & 0.0& 0.0 & 4.6\\
& RAP    & 67 & 76 & 74 & 48 & 17 & 9 & 51\\
& AutoHD   & \textbf{95.6} & \textbf{96.4} & \textbf{82.9} & \textbf{55.6} & 13.4 & 0.0 & \textbf{59.1}\\ \hline 

\end{tabular}
\end{table}

\subsection{Choice of Heuristic Functions}
In this subsection, we analyze the impact of heuristic function selection on performance. Specifically, we compare two strategies, including selecting the best heuristic functions across all generations and selecting the best heuristic functions from the final generation. Experiments are conducted using GPT 4o-mini. The results, presented in Table~\ref{tab:heuristic_choice}, indicate that LLMs explore various heuristic functions during the evolution process. Selecting the best heuristic function across all generations leads to more robust and stable performance compared to relying solely on the final generation. 

\begin{table}[t]
\centering
\caption{Ablation study on the choice of heuristic functions using GPT 4o-mini for Blocksworld, Game of 24, and Rubik's Cube.}
    \label{tab:heuristic_choice}
    \vspace{0.1cm}
\begin{tabular}{l|ccc}
 \toprule
 & \multicolumn{1}{c}{Blocksworld}                                  & \multirow{2}{*}{Game of 24} & \multirow{2}{*}{Rubik's cube} \\
 & Step 2                              &                               \\ \hline
All generations & \multicolumn{1}{c}{95.7} & 54 & 82.5 \\
Final generation  & \multicolumn{1}{c}{93.3} & 41 & 65.6 \\
\bottomrule
\end{tabular}
\end{table}

\subsection{Datasets}
\begin{figure}[t]
    \vspace{-0.0cm}
    \begin{center}
    \begin{tabular}{@{\hspace{0cm}}c@{\hspace{-0.0cm}}c}
    \includegraphics[width=0.5\linewidth]{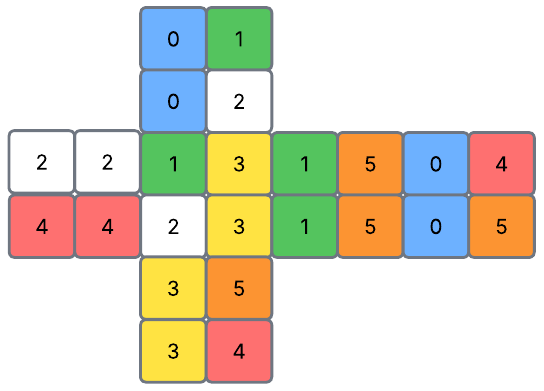} &
    \includegraphics[width=0.5\linewidth]{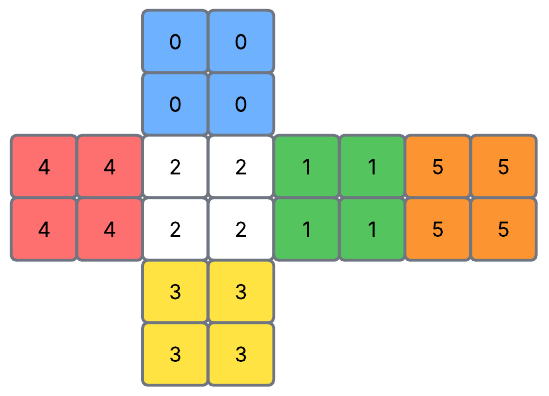} \\
    (a) a scrambled initial state  & (b) a goal state \\
    \end{tabular}
    \end{center}\vspace{-0.0cm}
    \caption{An example in Rubik's Cube dataset. The task is to transform a cube from a scrambled initial state to a goal state. }
     \label{fig:cube}
     \vspace{-0.0cm}
\end{figure}

Rubik’s Cube~\citep{ding2023everything} is a well-known puzzle-solving benchmark requiring multi-step spatial planning. The cube consists of six faces, each divided into four smaller squares, with each square assigned one of six distinct colors. The objective is to manipulate the cube from an initial scrambled state to a solved state, where each face is uniformly colored. This is accomplished through a sequence of predefined rotational moves applied to individual layers of the cube. Figure~\ref{fig:cube} provides a visual representation of the Rubik’s Cube dataset used in this work. The left subfigure shows a scrambled cube configuration, representing an initial state in the dataset. The arrangement of colored squares encodes the current state of the puzzle, which requires a sequence of moves to reach the solved state. The right subfigure shows a goal state, where each face of the cube is uniformly colored. The dataset consists of cube states that are at most four moves away from the solved configuration. 

\section{Search Algorithm}
In this section, we present the details of the search algorithms, specifically Greedy BFS and $A^*$, as outlined in Algorithm~\ref{alg:bfs} and Algorithm~\ref{alg:astar}, respectively. 

\begin{algorithm}
    \caption{Greedy BFS}\label{alg:bfs}
\begin{algorithmic}
\STATE {\bfseries Require:} Initial state $s_0$, heuristic function $H(\cdot)$, action space $\mathcal{A}_\theta(\cdot)$, transition function $T_\theta(\cdot)$
\STATE $Q \leftarrow \{s_0\}$
\WHILE {Q is not empty}
\STATE $s \leftarrow \text{argmax}_{s' \in Q} H(s')$ 
\STATE $Q \leftarrow Q \setminus \{s\}$
\FOR{$a \in \mathcal{A}_\theta(s)$}
\STATE $s' \leftarrow T_{\theta}(s,a)$
\STATE Return if $s'$ is a goal state 
\STATE $Q \leftarrow Q \cup \{s'\}$
\ENDFOR
\ENDWHILE
\end{algorithmic}
\end{algorithm}

\begin{algorithm}
    \caption{$\text{A}^*$ Search}\label{alg:astar}
\begin{algorithmic}
\STATE {\bfseries Require:} Initial state $s_0$, heuristic function $H(\cdot)$, cost function $G(\cdot)$, action space $\mathcal{A}_\theta(\cdot)$, transition function $T_\theta(\cdot)$
\STATE $Q \leftarrow \{s_0\}$
\WHILE {Q is not empty}
\STATE $s \leftarrow \text{argmax}_{s' \in Q} H(s') + G(s')$ 
\STATE $Q \leftarrow Q \setminus \{s\}$
\FOR{$a \in \mathcal{A}_\theta(s)$}
\STATE $s' \leftarrow T_{\theta}(s,a)$
\STATE Return if $s'$ is a goal state 
\STATE $Q \leftarrow Q \cup \{s'\}$
\ENDFOR
\ENDWHILE
\end{algorithmic}
\end{algorithm}

\section{Heuristic Functions}


In this section, we show some examples of heuristic functions generated by the LLMs.  Table~\ref{tab:heuristic_descriptions} shows some three representative examples generated by GPT 4o-mini. In the Blocksworld, the heuristic function estimates the effort required to transform an initial block configuration into a target configuration. It operates by first identifying the number of misplaced blocks that are not in their correct positions in the goal state. Additionally, it accounts for the cumulative positional difference, which measures how far each misplaced block is from its correct position. The final heuristic value is computed as the sum of these two terms. For the Game of 24, the heuristic function evaluates the proximity of a given set of numbers to the target value of 24. Given an input list of numbers, the heuristic iterates through all possible permutations of the numbers and arithmetic operations. The heuristic value is defined as the smallest absolute difference between 24 and the computed results of these expressions. This approach ensures that the heuristic effectively captures the validity of forming an expression that reaches 24. For the $2\times2$ Rubik’s Cube, the heuristic function estimates the number of moves required to reach a solved state by examining the uniformity of the cube's faces. Specifically, it counts the number of faces where all four squares are of the same color. Since a completely solved cube has six uniform faces, the heuristic value is computed as the total number of faces, i.e. six, minus the count of uniform faces. This function serves as a coarse but effective estimate of the cube’s disorder, guiding the search process toward configurations with increasing face uniformity. These examples illustrate the capacity of LLMs to generate domain-specific heuristic functions that align with problem-solving strategies typically designed by human experts.

\begin{table}[t]
\centering
\caption{Example heuristic functions generated by GPT 4o-mini.}
\label{tab:heuristic_descriptions}
\begin{tabular}{>{\centering\arraybackslash}m{1.3cm}>{\centering\arraybackslash}m{8.1cm}>{\centering\arraybackslash}m{3.2cm}}
\toprule
             & Heuristic Functions & Explanations \\ \midrule
Blocksworld  &    \includegraphics[width=1.0\linewidth]{FIG/heuristic_example.pdf}                 &   \begin{tabular}[c]{@{}l@{}} The heuristic function \\ computes a heuristic \\value by summing the \\ number of misplaced \\ blocks and their \\ cumulative positional \\ differences.\\  \end{tabular}         \\ \midrule
Game of 24   &       \includegraphics[width=1.0\linewidth]{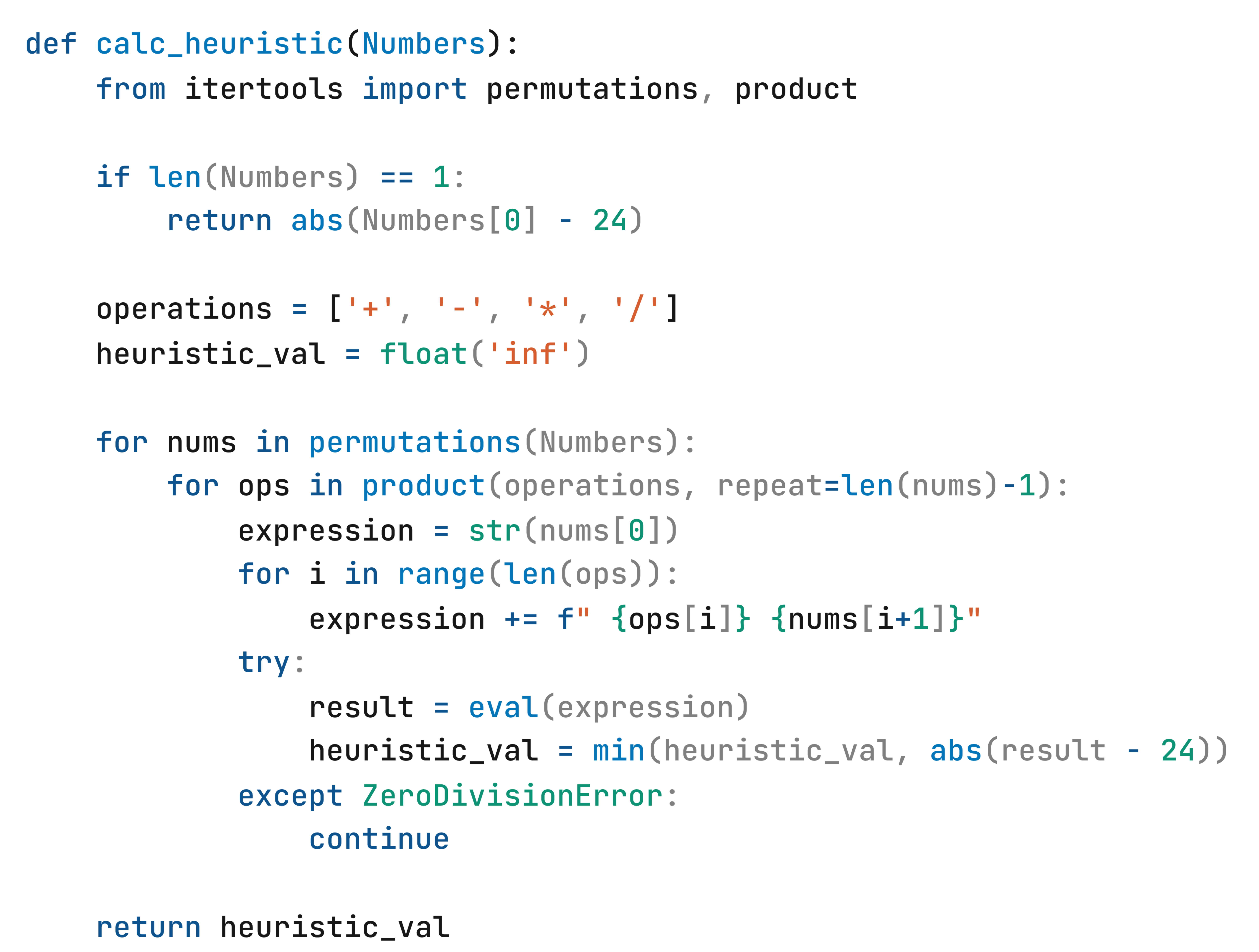}           &   \begin{tabular}[c]{@{}l@{}}  The function calculates \\ the  smallest absolute \\difference  between 24 \\ and the results of all \\ possible arithmetic \\ expressions formed \\ using a given list of \\ numbers and \\ operations.   \end{tabular}        \\ \midrule
Rubik's Cube &    \includegraphics[width=1.0\linewidth]{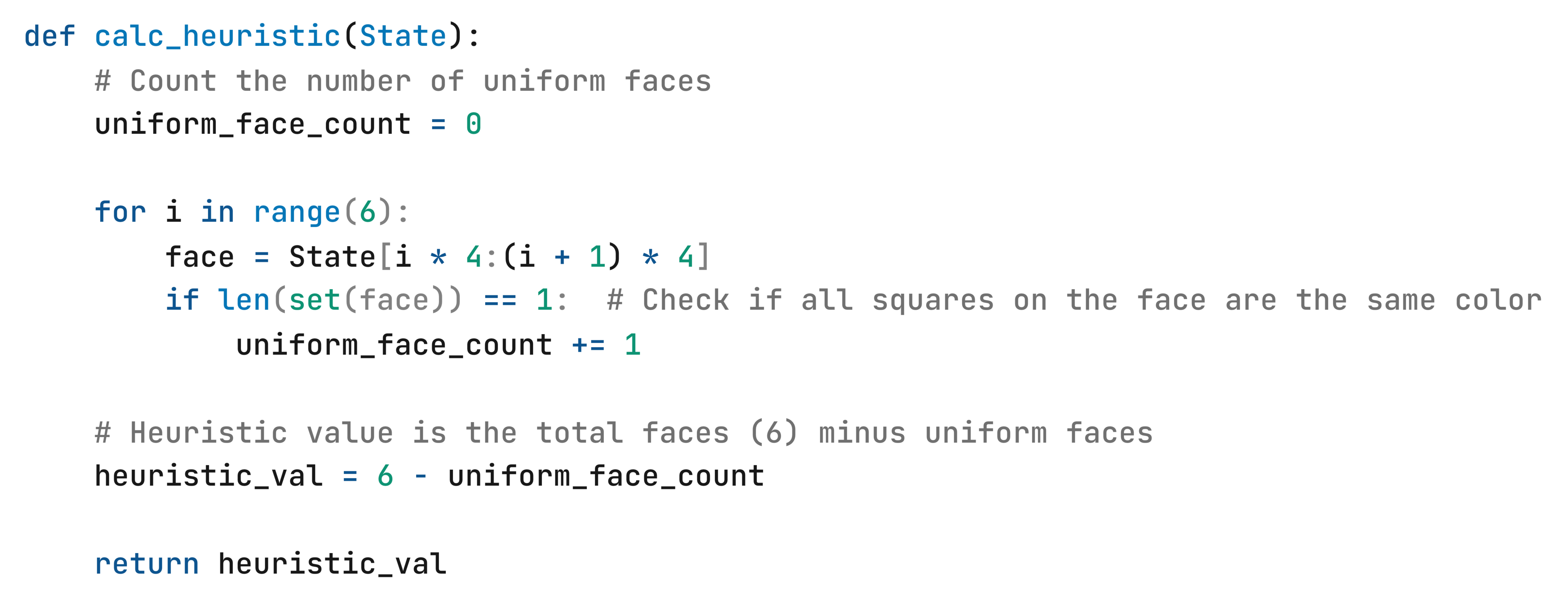}                  &  \begin{tabular}[c]{@{}l@{}}This heuristic function \\estimates the cost  to  \\ solve a $2\times2$  Rubik's\\ Cube by counting the \\ number of  non-uniform \\ faces on the  cube. \end{tabular}      
\\\bottomrule
\end{tabular}
\end{table}


\end{document}